\documentclass[10pt,journal,final]{IEEEtran}
\usepackage[latin9]{inputenc}
\usepackage{url}
\usepackage{amsmath}
\usepackage{amssymb}
\usepackage{graphicx}

\makeatletter

\newcommand{\lyxdot}{.}

\usepackage{url}
\usepackage{amsfonts}
\providecommand{\U}[1]{\protect\rule{.1in}{.1in}}

\makeatother

\begin{document}
\title{Practical Control for Multicopters to Avoid Non-Cooperative Moving
Obstacles}
\author{Quan~Quan, \emph{Member,~IEEE}, Rao Fu, and Kai-Yuan~Cai\thanks{Q. Quan, R. Fu and K-Y. Cai are with the School of Automation Science
and Electrical Engineering, Beihang University, Beijing 100191, China
(e-mail: qq\_buaa@buaa.edu.cn; buaafurao@buaa.edu.cn; kycai@buaa.edu.cn).}}
\maketitle
\begin{abstract}
Unmanned Aerial Vehicles (UAVs) are now becoming increasingly accessible
to amateur and commercial users alike. The main task for UAVs is to
keep a prescribed separation with obstacles in the air. In this paper,
a collision-avoidance control method for non-cooperative moving obstacles
is proposed for a multicopter with the altitude hold mode by using
a Lyapunov-like barrier function. Lyapunov-like functions are designed
elaborately, based on which formal analysis and proofs of the proposed
control are made to show that the collision-avoidance control problem
can be solved if the moving obstacle is slower than the multicopter.
The result can be extended to some cases of multiple obstacles. What
is more, by the proposed control, a multicopter can keep away from
obstacles as soon as possible, once obstacles enter into the safety
area of the multicopter accidentally, and converge to the waypoint.
Simulations and experiments are given to show the effectiveness of
the proposed method by showing the distance between UAV and waypoint,
obstacles respectively.
\end{abstract}


\section{Introduction}

\subsection{Background}

Airspace is utilized today by far lesser aircraft than it can accommodate,
especially low altitude airspace. There are more and more applications
for UAVs in low altitude airspace, ranging from on-demand package
delivery to traffic and wildlife surveillance, an inspection of infrastructure,
search and rescue, agriculture, and cinematography. Air traffic for
UAVs attracted more and more research \cite{IoD(2016)},\cite{Devasia(2016)}.
One of the chief concerns regarding UAVs is the possibility of collision
\cite{Jenie(2017)},\cite{Mihaela(2019)}. For such a purpose, a collision-avoidance
scheme must present a feasible path for the UAV to take in order to
maintain a minimum separation distance from obstacles in the air.
Traditionally, the main role of air traffic management (ATM) is to
keep a prescribed separation among all aircraft by using centralized
control. However, it is infeasible for increasing UAVs because the
traditional control method lacks scalability. In order to address
such a problem, free flight is a developing air traffic control method
that uses no centralized control \cite{Kuchar(2000)},\cite{Hoekstra2001}.
Free flight is very challenging in low altitude airspace because the
environment is complex and dynamic. Namely, many obstacles are dense,
moving \cite{Lin(2017)}, and unpredictable \cite{Jenie(2018)}.

\subsection{Exsiting Avoidance Techniques}

The avoidance technique has been studied extensively \cite{Huang(2019)},\cite{Mcfadyen(2016)}.
The close-range air avoidance algorithms can be roughly divided into
four categories \cite{Kuchar(2000)}: trajectory-projection based
method, online table based method, force field based method, and optimal
trajectory method. A simulation study of four typical collision avoidance
methods can be found in \cite{Mueller(2016)}.
\begin{itemize}
\item \textbf{Trajectory-Projection Based Method} 
\end{itemize}
The trajectory-based projection method needs to estimate the current
state of the obstacle (such as position and velocity) and predict
its trajectory. If the position of a UAV enters this dangerous area,
a new heading angle command is immediately generated to cause the
UAV to leave the dangerous area. After leaving the dangerous area,
the UAV returns back to the original trajectory along with the original
heading. The velocity obstacle method \cite{Fiorini(1998)} is one
of the classical methods. This method is simple, straightforward,
and easy to implement. In \cite{Thanh(2018)}, the collision avoidance
problem was solved for multicopters by combining geometric constraints
and kinematics equations. The maneuver generated from the selective
velocity obstacle method can avoid obstacles while incorporating the
right-of-way rules from the original route \cite{Jenie(2015)}. Furthermore,
a three-dimensional extension of the velocity obstacle method can
reactively generate an avoidance maneuver by changing the vehicle
velocity vector based on the encounter geometry \cite{Jenie(2016)}.
The trajectory projection method is more suitable for the control
avoidance strategy based on the vision sensor or\ the fixed-wing
aircraft maneuver. Because the visual sensor often only obtains the
orientation of the obstacle and cannot measure the relative position,
it is a relatively straightforward control method by changing the
flight heading angle \cite{Mcfadyen(2016)}. As for the fixed-wing
aircraft, the change in direction is better to slowing down the speed.
\begin{itemize}
\item \textbf{Online Table Based Method} 
\end{itemize}
Once the obstacles' status (location and speed) is received, this
method will search its online form and determine the best way to maneuver.
In \cite{Ong(2016)}, a robust and efficient algorithm was proposed
based on decomposing a large multi-agent Markov decision process and
integrating its solutions to generate recommendations for each aircraft.
In \cite{Saunders(2005)}, the concept of rapidly exploring random
trees (RRT) originally investigated in \cite{LaValle(2001)} was used
to find dynamically feasible obstacle-free paths. After obtaining
an obstacle-free path, a path planner was used to avoid obstacles.
When the environment is complex or local information is observed,
the online table based method may fail or requires larger computation
time to update frequently.
\begin{itemize}
\item \textbf{Force Field Based Method} 
\end{itemize}
Force field based methods typically use attractive forces (maintaining
on the original path or following the original destination) and repulsive
forces (avoiding potential conflicts) to generate control commands
for the next step. The weights of the different forces are adjusted
online to balance the trade-offs between the different forces. In
\cite{Balazs(2018)}, a general and distributed air traffic control
scheme using autonomous UAVs is proposed for dense traffic conditions,
and 30 autonomous UAVs are used for verification in coordinated outdoor
flight. In \cite{Viragh(2016)}, the traffic simulation scene of dense
multicopters in 2D and 3D open space was studied in the presence of
a real environment of sensor noise, communication delay, limited communication
range, limited sensor update rate, and limited power. One disadvantage
of the force field based method is the risk of getting stuck in local
minima \cite{Hernandez(2011)}.
\begin{itemize}
\item \textbf{Optimal Trajectory Method} 
\end{itemize}
As for the optimal trajectory method, it uses optimization methods
to plan feasible trajectories. In \cite{Boivin(2008)}, the predictive
control algorithm was used to calculate the optimal waypoint, which
can avoid the static obstacles detected on the way while reaching
the destination. In \cite{Yang(2018)}, a guidance method with obstacle
avoidance capability was designed and analyzed to help the aircraft
reach its destination quickly while avoiding collision with other
aircraft. This method expressed the collision avoidance problem as
a Markov decision process and solved it using the Monte Carlo tree
search method. Based on approximate aircraft dynamics, a mixed-integer
linear programming approach was used to create flight paths without
collisions \cite{Richards(2002)},\cite{Alonso(2010)}. A real-time
path planning algorithm was proposed for UAVs to avoid collisions
with other aircraft \cite{Lin(2015)}. The reachable set was used
to represent the set of possible trajectories of the obstacle aircraft
and was used for collision prediction in UAV path planning. The optimal
trajectory method is very suited for known static obstacles. As for
moving obstacle avoidance, the optimal trajectory method requires
to predict the motion of moving obstacles. What is more, this method
requires a larger computation time to update frequently.

\subsection{Proposed Force Field Method}

As for our problem, the force field method possesses a number of distinct
advantages when compared to other methods \cite{Hernandez(2011)},\cite{Yadollah(2017)}.
First, it uses a simple formula to react to obstacles as they appear,
which have low demand for computing resources, making them very suitable
for real-time avoidance. It can work on a processor with limited processing
capabilities. Moreover, this method can distribute onto each UAV.
If the UAV has an active detection device, then these UAVs need not
communicate with each other.

In this paper, a force field method is proposed for multicopters to
avoid multiple moving obstacles by using a Lyapunov-like barrier function.
Because of the larger uncertainties, the problem for multicopters
in the air differs from those for some indoor robots with a highly
accurate position estimation and control. The conflict of an obstacle
and a multicopter is often defined that their distance is less than
a safety distance. However, the conflict will happen in practice even
if conflict avoidance is proved formally because some assumptions
will be violated in practice. Even though the multicopters may not
have a real collision in physics because the safety distance is often
set largely by considering uncertainties. In most literature, if their
distance is less than a safety distance, then their control schemes
either do not work or even push the agent towards the center of the
safety area rather than the outside of the safety area. For example,
some studies have used the following barrier function terms for collision
avoidance, such as $1\left/\left(\left\Vert \mathbf{p}_{i}-\mathbf{p}_{j}\right\Vert -R\right)\right.$\cite{Quan(2017)}
or $\ln\left(\left\Vert \mathbf{p}_{i}-\mathbf{p}_{j}\right\Vert -R\right)$
\cite{Panagou(2016)}, where $\mathbf{p}_{i}$, $\mathbf{p}_{j}$
are two multicopters'{}positions, and $R>0$ is the separation distance.
The principle is to design a controller to make the barrier function
terms bounded so that $\left\Vert \mathbf{p}_{i}-\mathbf{p}_{j}\right\Vert >R$
if $\left\Vert \mathbf{p}_{i}\left(0\right)-\mathbf{p}_{j}\left(0\right)\right\Vert >R$.
Otherwise, $\left\Vert \mathbf{p}_{i}-\mathbf{p}_{j}\right\Vert =R$
will make the barrier function term unbounded. The separation distance
for robots indoor is often the sum of the two robots'{}physical radius,
namely $\left\Vert \mathbf{p}_{i}-\mathbf{p}_{j}\right\Vert <R$ will
not happen in practice. But, the separation distance is set largely
for multicopters compared with their sizes. Due to some uncertainties
such as communication delay, $\left\Vert \mathbf{p}_{i}-\mathbf{p}_{j}\right\Vert <R$
will happen in the air. As a consequence, the control corresponding
to the barrier function terms mentioned above will make $\left\Vert \mathbf{p}_{i}-\mathbf{p}_{j}\right\Vert \rightarrow0$
if $1\left/\left(\left\Vert \mathbf{p}_{i}-\mathbf{p}_{j}\right\Vert -R\right)\right.$
is used (the two multicopters are pushed together by the designed
controller) or appear numerical computation error if $\ln\left(\left\Vert \mathbf{p}_{i}-\mathbf{p}_{j}\right\Vert -R\right)$
is used.

\subsection{Contributions}

In view of this, a practical controller is proposed to solve the moving
non-cooperative obstacle avoidance problem, where the non-cooperative
obstacle is that will not make avoidance with the multicopter. Compared
with cooperative obstacles, the non-cooperative obstacle is more difficult
to deal with. Since the multicopter's speed is confined, it cannot
avoid some non-cooperative (hostile) moving obstacles such as a fast
bullet. However, cooperative obstacles and the multicopter can make
avoidance with each other by slowing speed or changing their directions
simultaneously\ \cite{David(2011)}. On the other hand, when obstacles
are moving, the convergence analysis is difficult because the resulting
equilibriums maybe not constant anymore, which are not easy to determine
and analyze. This is different from the analysis of static obstacles.
For this, in this paper, the instability about angle rather than position
is proved for the case that the multicopter is in front of an obstacle
moving direction. With this result, the convergence analysis is proved
for one moving obstacle avoidance control problem. Furthermore, one
moving obstacle avoidance control extends to two types of multiple
moving obstacle avoidance control problems. The contributions lie
in the practicability and properties of this method. The practicability
of the proposed control lies in the following two features: (i) a
double integral model with the given velocity command as input is
proposed for multicopters, where the maneuverability has been taken
into consideration; (ii) the maximum velocity command in the proposed
distributed controller is confined. With the two features, the properties
of the proposed method lie in the following two features: (i) formal
proofs about conflict avoidance for one moving obstacle problem are
given; (ii) formal proofs about the convergence to the desired waypoints
are further given.

\section{Problem Formulation}

In this section, a multicopter control model and an obstacle model
are introduced first, including two types of areas, namely safety
area and avoidance area, used for control. Then, the obstacle avoidance
control problem is formulated.

\subsection{Control Model}

\subsubsection{Multicopter\ Control Model}

Many organizations or companies have designed some open-source semi-autonomous
autopilots or offered semi-autonomous autopilots with Software Development
Kits. The semi-autonomous autopilots can be used for velocity control
of multicopters. For example, A3 autopilots released by DJI allow
the range of the horizontal velocity command from $-10$m/s$\sim10$m/s
\cite{A3}. With such an autopilot, the velocity of a multicopter
can track a given velocity command in a reasonable time. Not only
can this avoid the trouble of modifying the low-level source code
of autopilots, but also it can utilize commercial autopilots to complete
various tasks. Based on this, we suppose that there is a multicopter
with the altitude hold mode in local airspace satisfying the following
model
\begin{align}
\mathbf{\dot{p}} & =\mathbf{v}\nonumber \\
\mathbf{\dot{v}} & =-l\left(\mathbf{v}-\mathbf{v}_{\text{c}}\right)\label{positionmodel_ab_con_i}
\end{align}
where $l>0,$ $\mathbf{p}\in{{\mathbb{R}}^{2}}$ and $\mathbf{v}\in{{\mathbb{R}}^{2}}$
are the position and velocity of the multicopter{,} $\mathbf{v}_{\text{c}}\in{{\mathbb{R}}^{2}}$
is the velocity command of the multicopter. The control gain $l$
depends on the multicopter and the semi-autonomous autopilot used,
which can be obtained through flight experiments. From the model (\ref{positionmodel_ab_con_i}),
$\lim_{t\rightarrow\infty}\left\Vert \mathbf{v}\left(t\right)-\mathbf{v}_{\text{c}}\right\Vert =0$
if $\mathbf{v}_{\text{c}}$ is constant. Here, the velocity command
$\mathbf{v}_{\text{c}}$ for the multicopter is subject to a saturation
defined as 
\begin{equation}
\text{sa}{\text{t}}\left(\mathbf{v},{v_{\text{m}}}\right)\triangleq\left\{ \begin{array}{cc}
\mathbf{v} & \left\Vert \mathbf{v}\right\Vert \leq{v_{\text{m}}}\\
{v_{\text{m}}}\frac{\mathbf{v}}{\left\Vert \mathbf{v}\right\Vert } & \left\Vert \mathbf{v}\right\Vert >{v_{\text{m}}}
\end{array}\right.\label{saturation}
\end{equation}
where ${v_{\text{m}}>0}$ is the setting maximum speed of the multicopter,
$\mathbf{v}\triangleq\lbrack{{v}_{1}}$ ${{v}_{2}}]{^{\text{T}}}\in{{\mathbb{R}}^{2}}$.
This implies 
\begin{equation}
\left\Vert \mathbf{v}_{\text{c}}\right\Vert \leq{v_{\text{m}}.}\label{limit1}
\end{equation}
The saturation function sa${\text{t}}\left(\mathbf{v},{v_{\text{m}}}\right)$
and the vector $\mathbf{v}$ are parallel all the time so the multicopter
can keep the flying direction the same if $\left\Vert \mathbf{v}\right\Vert >{v_{\text{m}}}$
\cite{Quan(2017)}. The saturation function can be rewritten as 
\[
\text{sat}\left(\mathbf{v},v_{\text{m}}\right)=\kappa_{v_{\text{m}}}\left(\mathbf{v}\right)\mathbf{v}
\]
where 
\[
{{\kappa}_{{v_{\text{m}}}}}\left(\mathbf{v}\right)\triangleq\left\{ \begin{array}{c}
1,\\
\frac{{v_{\text{m}}}}{\left\Vert \mathbf{v}\right\Vert },
\end{array}\begin{array}{c}
\left\Vert \mathbf{v}\right\Vert \leq{v_{\text{m}}}\\
\left\Vert \mathbf{v}\right\Vert >{v_{\text{m}}}
\end{array}\right..
\]
It is obvious that $0<{{\kappa}_{{v_{\text{m}}}}}\left(\mathbf{v}\right)\leq1$.
Sometimes, ${{\kappa}_{{v_{\text{m}}}}}\left(\mathbf{v}\right)$ will
be written as ${{\kappa}_{{v_{\text{m}}}}}$ for short. According
to this, if and only if $\mathbf{v=0},$ then 
\[
\mathbf{v}^{\text{T}}\text{sa}{\text{t}}\left(\mathbf{v},{v_{\text{m}}}\right)=0.
\]
In the presence of the saturation constraint, the velocity is confined
no matter how the controller is designed.

\textbf{Proposition 1}. If $\left\Vert \mathbf{v}\left(0\right)\right\Vert \leq{v_{\text{m}}}$
and the model (\ref{positionmodel_ab_con_i}) is subject to (\ref{saturation}),
namely 
\begin{align*}
\mathbf{\dot{p}} & =\mathbf{v}\\
\mathbf{\dot{v}} & =-l\left(\mathbf{v}-\text{sa}{\text{t}}\left(\mathbf{v}_{\text{c}},{v_{\text{m}}}\right)\right)
\end{align*}
then $\left\Vert \mathbf{v}\left(t\right)\right\Vert \leq{v_{\text{m}},}$
$t\geq0{.}$

\textit{Proof}. See \emph{Appendix}. $\square$

\textbf{Remark 1}. Because the propellers often provide the unidirectional
upward thrust, the descending and climbing ability of a multicopter
are different, the model (\ref{positionmodel_ab_con_i}) is not very
suit for the altitude control. So, here, we only consider a multicopter
with the altitude hold mode in 2D space for simplicity. When a Vertical
TakeOff and Landing (VTOL) UAV takes flight with the altitude hold
mode, the model (\ref{positionmodel_ab_con_i}) can be adopted.

\subsubsection{Obstacle Model}

In the same local airspace, there exists a moving obstacle (it may
be an aircraft or a balloon) defined as 
\[
\mathcal{O}=\left\{ \mathbf{x}\in\mathbb{R}^{2}\left\vert \left\Vert \mathbf{x}-\mathbf{p}_{\text{o}}\right\Vert <r_{\text{o}}\right.\right\} 
\]
where $r_{\text{o}}>0$ is the obstacle radius, and $\mathbf{p}_{\text{o}}\in\mathbb{R}^{2}$
is the center of mass of the obstacle. Define 
\[
\boldsymbol{\xi}_{\text{o}}\triangleq\mathbf{p}_{\text{o}}+\frac{1}{l}\mathbf{v}_{\text{o}}
\]
where $\mathbf{v}_{\text{o}}\in\mathbb{R}^{2}$ is the velocity of
the obstacle. The obstacle satisfies the following model 
\[
\max\left\Vert \boldsymbol{\dot{\xi}}_{\text{o}}\right\Vert \leq v_{\text{o}}
\]
where $v_{\text{o}}>0.$ This is a general model for any obstacle
with bounded velocity and acceleration. Let 
\begin{equation}
\boldsymbol{\dot{\xi}}_{\text{o}}=\mathbf{a}_{\text{o}}.\label{ksi0}
\end{equation}
with $\max\left\Vert \mathbf{a}_{\text{o}}\right\Vert \leq v_{\text{o}}.$
Then (\ref{ksi0}) can be rewritten as 
\begin{align}
\mathbf{\dot{p}}_{\text{o}} & =\mathbf{v}_{\text{o}}\nonumber \\
\mathbf{\dot{v}}_{\text{o}} & =-l\left(\mathbf{v}_{\text{o}}-\mathbf{a}_{\text{o}}\right).\label{obstacle}
\end{align}
In particular, if $\left\Vert \mathbf{v}_{\text{o}}\left(0\right)\right\Vert \leq v_{\text{o}}\ $and
$\mathbf{a}_{\text{o}}=\mathbf{v}_{\text{o}}\left(0\right),$ then
the obstacle is moving with a constant velocity. If $\mathbf{v}{_{\text{o}}=\mathbf{0},}$
then the obstacle is static.

\subsubsection{Filtered Position Model}

\begin{figure}[h]
\begin{centering}
\includegraphics{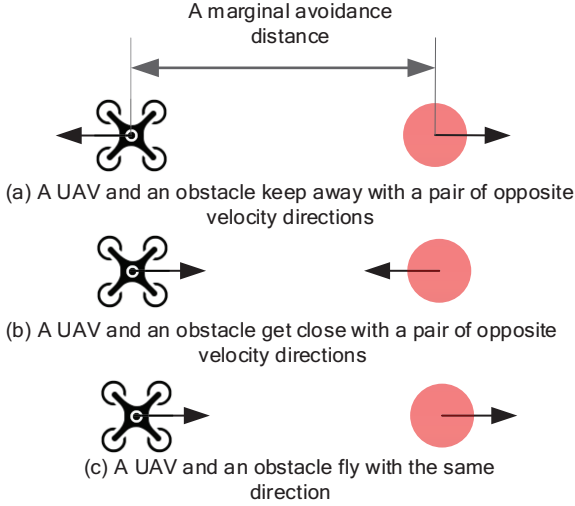} 
\par\end{centering}
\caption{Intuitive interpretation for filtered position.}
\label{Intuitive}
\end{figure}

As shown in Fig. \ref{Intuitive}, although the position distances
of the three cases are the same, namely a marginal avoidance distance,
the case in Fig. \ref{Intuitive}(b) needs to carry out avoidance
urgently by considering the velocity. However, the case in Fig. \ref{Intuitive}(a)
in fact does not need to be considered. 

With such an intuition, a filtered position is defined as follows:
\begin{equation}
\boldsymbol{\xi}\triangleq{\mathbf{p}}+\frac{1}{{l}}\mathbf{v}.\label{FilteredPosition}
\end{equation}
{Then} 
\begin{align}
\boldsymbol{\dot{\xi}} & =\mathbf{\dot{p}}+\frac{1}{{l}}\mathbf{\dot{v}}\nonumber \\
 & =\mathbf{v}-\frac{1}{{l}}{l}\left(\mathbf{v}-\mathbf{v}_{\text{c}}\right)\nonumber \\
 & =\mathbf{v}_{\text{c}}.\label{filteredposdyn}
\end{align}
Define 
\begin{align}
\mathbf{\tilde{p}}{_{\text{o}}} & \triangleq\mathbf{p}-{{\mathbf{p}}_{\text{o}}}\nonumber \\
\mathbf{\tilde{v}}{_{\text{o}}} & \triangleq\mathbf{v}-{{\mathbf{v}}_{\text{o}}}\nonumber \\
\boldsymbol{\tilde{\xi}}{_{\text{o}}} & \triangleq\mathbf{\tilde{p}}{_{\text{o}}}+\frac{1}{{l}}\mathbf{\tilde{v}}{_{\text{o}}}\label{errors}
\end{align}
{and} 
\begin{equation}
r_{\text{v}}=\frac{{v_{\text{m}}+}v_{\text{o}}}{{l}}.\label{rv}
\end{equation}
According to \textit{Proposition 1}, we have 
\begin{equation}
\frac{1}{{l}}\left\Vert \mathbf{\tilde{v}}{_{\text{o}}}\right\Vert \leq r_{\text{v}}.\label{rv1}
\end{equation}
In the following, a relationship between the position error and the
filtered position error is shown.

\textbf{Proposition 2}. For the multicopter and the obstacle, \emph{if
and only if} the filtered position error satisfies 
\begin{equation}
\left\Vert \boldsymbol{\tilde{\xi}}{_{\text{o}}}\left(t\right)\right\Vert \geq\sqrt{r^{2}+r_{\text{v}}^{2}},\label{p3condition}
\end{equation}
and $\left\Vert \mathbf{\tilde{p}}{_{\text{o}}}\left(0\right)\right\Vert \geq r,$
then $\left\Vert \mathbf{\tilde{p}}{_{\text{o}}}\left(t\right)\right\Vert \geq r,$
where $t>0$. The relationship ``$=$''\ holds if $\frac{\mathbf{v}^{\text{T}}{{\mathbf{v}}_{\text{o}}}}{\left\Vert \mathbf{v}\right\Vert \left\Vert {{\mathbf{v}}_{\text{o}}}\right\Vert }=-1.$
Furthermore, if $\left\Vert \boldsymbol{\tilde{\xi}}{_{\text{o}}}\left(t\right)\right\Vert >\sqrt{r^{2}+r_{\text{v}}^{2}}$
and $\left\Vert \mathbf{\tilde{p}}{_{\text{o}}}\left(0\right)\right\Vert >r,$
then $\left\Vert \mathbf{\tilde{p}}{_{\text{o}}}\left(t\right)\right\Vert >r,$
where $t>0$.

\textit{Proof}. See \emph{Appendix}. $\square$

\subsection{Two Types of Areas around a Multicopter}

Two types of areas used for control, namely \emph{safety area} and
\emph{avoidance area}, are introduced. 
\begin{figure}[h]
\begin{centering}
\includegraphics[scale=0.6]{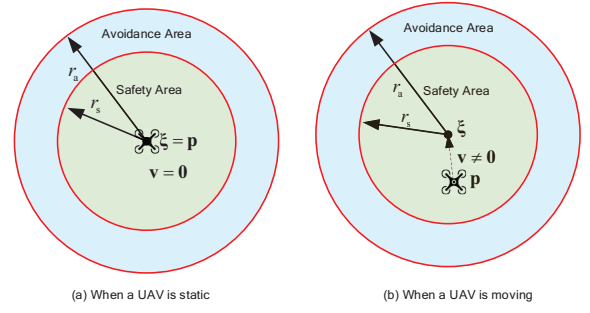} 
\par\end{centering}
\caption{Safety area and avoidance area of a UAV.}
\label{Threeaera}
\end{figure}

\subsubsection{Safety Area}

In order to avoid a conflict, as shown in Figure \ref{Threeaera},
the \emph{safety area} of a multicopter is defined as 
\begin{equation}
\mathcal{S}=\left\{ \mathbf{x}\in{{\mathbb{R}}^{2}}\left\vert \left\Vert \mathbf{x}-\boldsymbol{\xi}\right\Vert \leq r_{\text{s}}\right.\right\} \label{Safetyaera}
\end{equation}
where $r_{\text{s}}>0$ is the safety radius. It should be noted that
we consider the velocity of the multicopter in the definition of $\mathcal{S}$.
For the multicopter, no \emph{conflict }with the obstacle implies
\[
\mathcal{S}\cap\mathcal{O}=\varnothing
\]
namely 
\begin{equation}
\left\Vert \boldsymbol{\tilde{\xi}}{_{\text{o}}}\right\Vert >r_{\text{s}}+r_{\text{o}}.\label{sdis}
\end{equation}
\textit{Proposition 2} implies that the multicopter and the obstacle
will be separated largely enough if (\ref{sdis}) is satisfied with
a safety radius\emph{ }$r_{\text{s}}$ large enough.

\subsubsection{Avoidance Area}

Besides the safety area, there exists an \emph{avoidance area} used
for starting avoidance control. If the obstacle is out of the avoidance
area of the multicopter, then the obstacle will not need to be avoided.
For the multicopter, the \emph{avoidance area }for other multicopters
is\emph{ }defined as 
\begin{equation}
\mathcal{A}=\left\{ {\mathbf{x}}\in{{\mathbb{R}}^{2}}\left\vert \left\Vert \mathbf{x}-\boldsymbol{\xi}\right\Vert \leq r_{\text{a}}\right.\right\} \label{Avoidancearea}
\end{equation}
where $r_{\text{a}}>0$ is the \emph{avoidance radius}. It should
be noted that we consider the velocity of the multicopter in the definition
of $\mathcal{A}$. If 
\[
\mathcal{A}\cap\mathcal{O}\neq\varnothing,
\]
namely 
\[
\left\Vert \boldsymbol{\tilde{\xi}}{_{\text{o}}}\right\Vert \leq r_{\text{a}}
\]
then the obstacle should be avoided by the multicopter. When the obstacle
just enters into the avoidance area of the multicopter, it is required
that they have not conflicted at the beginning. Therefore,\textbf{
}we require 
\[
r_{\text{a}}>r_{\text{s}}.
\]

\subsection{Moving Obstacle Avoidance Control Problem}

Before introducing the problem, the following assumptions are imposed.

\textbf{Assumption 1}. The information of obstacle $\mathcal{O}$
can be detected, where the velocity of the obstacle $\mathbf{v}{_{\text{o}}}\ $is
constant.

\textbf{Assumption 2}. The multicopter's initial filtered position
error $\boldsymbol{\tilde{\xi}}{_{\text{o}}}\left(0\right)\in{{\mathbb{R}}^{2}}$
satisfies 
\[
\left\Vert \boldsymbol{\tilde{\xi}}{_{\text{o}}}\left(0\right)\right\Vert >r_{\text{s}}+{{r}_{\text{o}}}
\]
and $\left\Vert \mathbf{v}\left(0\right)\right\Vert \leq{v_{\text{m}}.}$

\textbf{Assumption 3}. The goal waypoint is static. Furthermore, if
the obstacle is static, then 
\[
\left\Vert \mathbf{p}_{\text{wp}}-\mathbf{p}_{\text{o}}\right\Vert \geq{r}_{\text{a}}+{{r}_{\text{o}}.}
\]

Based on \textit{Assumptions 1-3}, we have the \emph{moving} \emph{obstacle
avoidance control problem} stated in the following.

\textbf{Moving Obstacle} \textbf{Avoidance Control Problem}. Let $\mathbf{p}\in{{\mathbb{R}}^{2}}$
be the position of the multicopter, and ${{\mathbf{p}}_{\text{wp}}}\in{{\mathbb{R}}^{2}}$
be the goal waypoint. Under \textit{Assumptions 1-3}, design the velocity
input $\mathbf{v}_{\text{c}}$ for the multicopter modeled in (\ref{positionmodel_ab_con_i})
to guide it flying until it arrives at the goal waypoint ${{\mathbf{p}}_{\text{wp}}}$,
meanwhile avoiding colliding the obstacle $\mathcal{O}$. The one
moving obstacle avoidance control problem will be extended for the
multiple moving obstacles avoidance control problem.

\textbf{Remark 2}. With the help of surveillance systems on the ground
or detection devices onboard (cameras or radars), the information
of obstacle $\mathcal{O}$ can be detected. The assumption about the
constant obstacle velocity$\ $is only for the convenience of the
convergence proof to the waypoint. As pointed by the following \textit{Lemma
2}, the collision-avoidance control problem can be solved if the moving
obstacle is slower than the multicopter with only \textit{Assumption
2. Assumptions 2-3} imply that the multicopter and its waypoint are
not close to the obstacle too much initially. As for \textit{Assumption
3}, if ${r}_{\text{a}}=3$m, {it only requires that the waypoint
keeps away from the obstacle 3m. It is reasonable in practice.}

\section{Preliminaries}

\subsection{Line Integral Lyapunov Function}

In the following, we will design a new type of Lyapunov Functions,
called \emph{Line Integral Lyapunov Function. }This type of Lyapunov
functions is inspired by its scalar form \cite[p.74]{Slotine(1991)}.
If $xf\left(x\right)>0\ $for $x\neq0,$ then $V_{\text{li}}^{\prime}\left(y\right)=\int_{0}^{y}f\left(x\right)$d$x>0$
when $y\neq0.$ The derivative is $\dot{V}_{\text{li}}^{\prime}=f\left(y\right)\dot{y}.$
A line integral Lyapunov function for vectors is defined as 
\begin{equation}
V_{\text{li}}\left(\mathbf{y}\right)=\int_{C_{\mathbf{y}}}\text{sa}{\text{t}}\left(\mathbf{x},a\right)^{\text{T}}\text{d}\mathbf{x}\label{Vli}
\end{equation}
where $a>0,$ $\mathbf{x}\in\mathbf{
\mathbb{R}
}^{n},$ $C_{\mathbf{y}}$ is a line from $\mathbf{0}$ to $\mathbf{y}\in
\mathbb{R}
^{n}\mathbf{.}$ In the following lemma, we will show its properties.

\textbf{Lemma 1}. Suppose that the line integral Lyapunov function
$V_{\text{li}}$ is defined as (\ref{Vli}). Then (i) $V_{\text{li}}\left(\mathbf{y}\right)>0$
if $\left\Vert \mathbf{y}\right\Vert \neq0$; (ii) if $\left\Vert \mathbf{y}\right\Vert \rightarrow\infty,$
then $V_{\text{li}}\left(\mathbf{y}\right)\rightarrow\infty;$ (iii)
if $V_{\text{li}}\left(\mathbf{y}\right)$ is bounded, then $\left\Vert \mathbf{y}\right\Vert $
is bounded.

\textit{Proof}. Since 
\[
\text{sat}\left(\mathbf{x},a\right)=\kappa_{a}\left(\mathbf{x}\right)\mathbf{x}
\]
the function (\ref{Vli}) can be written as 
\begin{equation}
V_{\text{li}}\left(\mathbf{y}\right)=\int_{C_{\mathbf{y}}}{{\kappa}_{{a}}}\left(\mathbf{x}\right)\mathbf{x}^{\text{T}}\text{d}\mathbf{x}\label{Vli1}
\end{equation}
where 
\[
{{\kappa}_{{a}}}\left(\mathbf{x}\right)\triangleq\left\{ \begin{array}{c}
1,\\
\frac{{a}}{\left\Vert \mathbf{x}\right\Vert },
\end{array}\begin{array}{c}
\left\Vert \mathbf{x}\right\Vert \leq{a}\\
\left\Vert \mathbf{x}\right\Vert >{a}
\end{array}\right..
\]
Let $z=\left\Vert \mathbf{x}\right\Vert .$ Then the function (\ref{Vli1})
becomes 
\begin{align*}
V_{\text{li}}\left(\mathbf{y}\right) & =\int_{C_{\mathbf{y}}}\frac{{{\kappa}_{{a}}}\left(\mathbf{x}\right)}{2}\text{d}z^{2}\\
 & =\int_{0}^{\left\Vert \mathbf{y}\right\Vert }{{\kappa}_{{a}}}\left(\mathbf{x}\right)z\text{d}z.
\end{align*}

\begin{itemize}
\item If $\left\Vert \mathbf{y}\right\Vert \leq{a,}$ then ${{\kappa}_{{a}}}\left(\mathbf{x}\right)=1.$
Consequently, 
\begin{equation}
V_{\text{li}}\left(\mathbf{y}\right)=\frac{1}{2}\left\Vert \mathbf{y}\right\Vert ^{2}.\label{Vli2}
\end{equation}
\item If $\left\Vert \mathbf{y}\right\Vert >{a,}$ then 
\[
\int_{0}^{\left\Vert \mathbf{y}\right\Vert }{{\kappa}_{{a}}}\left(\mathbf{x}\right)z\text{d}z=\int_{0}^{a}z\text{d}z+\int_{a}^{\left\Vert \mathbf{y}\right\Vert }\frac{{a}}{\left\Vert \mathbf{x}\right\Vert }z\text{d}z.
\]
Since $z=\left\Vert \mathbf{x}\right\Vert ,$ we have 
\begin{equation}
V_{\text{li}}\left(\mathbf{y}\right)=\frac{1}{2}a^{2}+{a}\left(\left\Vert \mathbf{y}\right\Vert -a\right).\label{Vli3}
\end{equation}
\end{itemize}
Therefore, from the form of (\ref{Vli2}) and (\ref{Vli3}), we have
(i) $V_{\text{li}}\left(\mathbf{y}\right)>0$ if $\left\Vert \mathbf{y}\right\Vert \neq0$.
(ii) if $\left\Vert \mathbf{y}\right\Vert \rightarrow\infty,$ then
$V_{\text{li}}\left(\mathbf{y}\right)\rightarrow\infty;$ (iii) if
$V_{\text{li}}\left(\mathbf{y}\right)$ is bounded, then $\left\Vert \mathbf{y}\right\Vert $
is bounded. $\square$

\subsection{Two Smooth Functions}

\begin{figure}[h]
\begin{centering}
\includegraphics[scale=0.7]{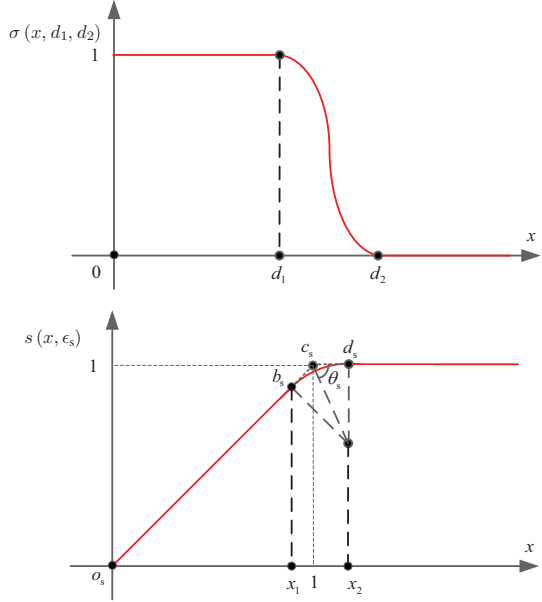} 
\par\end{centering}
\caption{Two smooth functions. For a smooth saturation function, $\theta_{\text{s}}=67.5^{\circ}.$}
\label{saturationa}
\end{figure}

Two smooth functions are defined for the following Lyapunov-like function
design. As shown in Figure \ref{saturationa}(a), define a second-order
differentiable `bump' function as \cite{Panagou(2016)} 
\begin{equation}
\sigma\left(x,d_{1},d_{2}\right)=\left\{ \begin{array}{c}
1\\
Ax^{3}+Bx^{2}+Cx+D\\
0
\end{array}\right.\begin{array}{c}
\text{if}\\
\text{if}\\
\text{if}
\end{array}\begin{array}{c}
x\leq d_{1}\\
d_{1}\leq x\leq d_{2}\\
d_{2}\leq x
\end{array}\label{zerofunction}
\end{equation}
with $A=-2\left/\left(d_{1}-d_{2}\right)^{3}\right.,$ $B=3\left(d_{1}+d_{2}\right)\left/\left(d_{1}-d_{2}\right)^{3}\right.,$
$C=-6d_{1}d_{2}\left/\left(d_{1}-d_{2}\right)^{3}\right.$ and $D=d_{2}^{2}\left(3d_{1}-d_{2}\right)\left/\left(d_{1}-d_{2}\right)^{3}\right.$.
The derivative of $\sigma\left(x,d_{1},d_{2}\right)$ with respect
to $x$ is 
\[
\frac{\partial\sigma\left(x,d_{1},d_{2}\right)}{\partial x}=\left\{ \begin{array}{c}
0\\
3Ax^{2}+2Bx+C\\
0
\end{array}\right.\begin{array}{c}
\text{if}\\
\text{if}\\
\text{if}
\end{array}\begin{array}{c}
x\leq d_{1}\\
d_{1}\leq x\leq d_{2}\\
d_{2}\leq x
\end{array}.
\]
Define another smooth function as shown in Figure \ref{saturationa}(b)
to approximate a saturation function 
\[
\bar{s}\left(x\right)=\min\left(x,1\right),x\geq0
\]
that 
\begin{equation}
s\left(x,\epsilon_{\text{s}}\right)=\left\{ \begin{array}{c}
x\\
\left(1-\epsilon_{\text{s}}\right)+\sqrt{\epsilon_{\text{s}}^{2}-\left(x-x_{2}\right)^{2}}\\
1
\end{array}\right.\begin{array}{c}
0\leq x\leq x_{1}\\
x_{1}\leq x\leq x_{2}\\
x_{2}\leq x
\end{array}\label{sat}
\end{equation}
with $x_{2}=1+\frac{1}{\tan67.5^{\circ}}\epsilon_{\text{s}}$ and
$x_{1}=x_{2}-\sin45^{\circ}\epsilon_{\text{s}}.$ Since it is required
$x_{1}\geq0$, one has $\epsilon_{\text{s}}\leq\frac{\tan67.5^{\circ}}{\tan67.5^{\circ}\sin45^{\circ}-1}.$
For any $\epsilon_{\text{s}}\in\left[{0,}\frac{\tan67.5^{\circ}}{\tan67.5^{\circ}\sin45^{\circ}-1}\right],$
it is easy to see 
\begin{equation}
s\left(x,\epsilon_{\text{s}}\right)\leq\bar{s}\left(x\right)\label{satinequ}
\end{equation}
and 
\begin{equation}
\lim_{\epsilon_{\text{s}}\rightarrow0}\underset{x\geq0}{\sup}\left\vert \bar{s}\left(x\right)-s\left(x,\epsilon_{\text{s}}\right)\right\vert =0.\label{sata}
\end{equation}
The derivative of $s\left(x,\epsilon_{\text{s}}\right)$ with respect
to $x$ is 
\[
\frac{\partial s\left(x,\epsilon_{\text{s}}\right)}{\partial x}=\left\{ \begin{array}{c}
1\\
\frac{x_{2}-x}{\sqrt{\epsilon_{\text{s}}^{2}-\left(x-x_{2}\right)^{2}}}\\
0
\end{array}\right.\begin{array}{c}
0\leq x\leq x_{1}\\
x_{1}\leq x\leq x_{2}\\
x_{2}\leq x
\end{array}.
\]
For any $\epsilon_{\text{s}}>0,$ we have $\underset{x\geq0}{\sup}\left\vert \partial s\left(x,\epsilon_{\text{s}}\right)\left/\partial x\right.\right\vert \leq1.$

\section{Obstacle Avoidance Control}

The idea of the proposed method is similar to that of the artificial
potential field (APF) method. In this method, the airspace is formulated
as an APF. For a given multicopter, only is the corresponding waypoint
assigned \textit{attractive potential}, while the obstacle is assigned
\textit{repulsive potentials}. A multicopter in the field will be
attracted to the waypoint, while being repelled by the obstacle.

\subsection{Error Model}

Define position error as 
\[
\mathbf{\tilde{p}}{_{\text{wp}}}\triangleq\mathbf{p}-{{\mathbf{p}}_{\text{wp}}}
\]
and the filtered position error as 
\[
\boldsymbol{\tilde{\xi}}{_{\text{wp}}}\triangleq\boldsymbol{\xi}-{{\mathbf{p}}_{\text{wp}}.}
\]
By (\ref{obstacle}) and (\ref{filteredposdyn}), the derivative of
the filtered errors are 
\begin{align}
\boldsymbol{\dot{\tilde{\xi}}}{_{\text{wp}}} & =\mathbf{v}_{\text{c}}\label{wpmodel}\\
\boldsymbol{\dot{\tilde{\xi}}}{_{\text{o}}} & =\mathbf{v}_{\text{c}}-\mathbf{a}_{\text{o}}.\label{obmodel}
\end{align}

\subsection{Lyapunov-Like Function Design and Analysis}

Define a smooth curve $C_{\boldsymbol{\tilde{\xi}}{_{\text{wp}}}}$
from $\mathbf{0}$ to $\boldsymbol{\tilde{\xi}}{_{\text{wp}}}$. Then,
the line integral of sa${\text{t}}\left(\boldsymbol{\tilde{\xi}}{_{\text{wp}}},{v_{\text{m}}}\right)$
along $C_{\boldsymbol{\tilde{\xi}}{_{\text{wp}}}}$ is 
\begin{equation}
V_{\text{w}}\left(\boldsymbol{\tilde{\xi}}{_{\text{wp}}}\right)=\int_{C_{\boldsymbol{\tilde{\xi}}{_{\text{wp}}}}}\text{sa}{\text{t}}\left(k_{1}\mathbf{x},v_{\text{m}}\right)^{\text{T}}\text{d}\mathbf{x}\label{Vw}
\end{equation}
where $k_{1}>0$. From the definition and \textit{Lemma 1}, $V_{\text{w}}\geq0.$
With the two defined functions (\ref{zerofunction}) and (\ref{sat}),
a Lyapunov-like function is defined as 
\begin{equation}
V_{\text{o}}\left(\left\Vert \boldsymbol{\tilde{\xi}}_{\text{o}}\right\Vert \right)=\frac{k_{2}\sigma_{\text{\text{o}}}\left(\left\Vert \boldsymbol{\tilde{\xi}}_{\text{o}}\right\Vert \right)}{\left(1+\epsilon\right)\left\Vert \boldsymbol{\tilde{\xi}}_{\text{o}}\right\Vert -\left(\gamma r_{\text{s}}+r_{\text{o}}\right)s\left(\frac{\left\Vert \boldsymbol{\tilde{\xi}}{}_{\text{o}}\right\Vert }{\gamma r_{\text{s}}+r_{\text{o}}},\epsilon_{\text{s}}\right)}\label{Vo}
\end{equation}
where $k_{2}{,}\epsilon_{\text{s}}>0,$ $\gamma>1,$ and $\sigma_{\text{o}}\left(x\right)\triangleq\sigma\left(x,\gamma r_{\text{s}}+{{r}_{\text{o}}},r_{\text{a}}\right).$
The function $V_{\text{o}}$ has the following properties:
\begin{itemize}
\item Property (i). $\partial V_{\text{o}}\left/\partial\left\Vert \boldsymbol{\tilde{\xi}}{_{\text{o}}}\right\Vert \right.\leq0$
as $V_{\text{o}}\ $is a nonincreasing function with respect to $\left\Vert \boldsymbol{\tilde{\xi}}{_{\text{o}}}\right\Vert $;
\item Property (ii). if $\left\Vert \boldsymbol{\tilde{\xi}}{_{\text{o}}}\right\Vert \geq r_{\text{a}}{,}$
namely the obstacle is out of the avoidance area of the multicopter,
then $\sigma_{_{\text{o}}}\left(\left\Vert \boldsymbol{\tilde{\xi}}{_{\text{o}}}\right\Vert \right)=0;$
consequently, $V_{\text{o}}=0$ and $\partial V_{\text{o}}\left/\partial\left\Vert \boldsymbol{\tilde{\xi}}{_{\text{o}}}\right\Vert \right.=0$;
on the other hand, if $V_{\text{o}}=0,$ then $\left\Vert \boldsymbol{\tilde{\xi}}{_{\text{o}}}\right\Vert \geq r_{\text{a}}{;}$
\item Property (iii). if $0<\left\Vert \boldsymbol{\tilde{\xi}}{_{\text{o}}}\right\Vert <\gamma r_{\text{s}}+{{r}_{\text{o}},}$
namely the safety area of the multicopter and the obstacle area are
close or have an intersection, then
\[
\sigma_{_{\text{o}}}\left(\left\Vert \boldsymbol{\tilde{\xi}}{_{\text{o}}}\right\Vert \right)=1
\]
and there exists a sufficiently small $\epsilon_{\text{s}}>0$ such
that 
\[
s\left(\frac{\left\Vert \boldsymbol{\tilde{\xi}}{_{\text{o}}}\right\Vert }{\gamma r_{\text{s}}+{{r}_{\text{o}}}},\epsilon_{\text{s}}\right)\approx\frac{\left\Vert \boldsymbol{\tilde{\xi}}{_{\text{o}}}\right\Vert }{\gamma r_{\text{s}}+{{r}_{\text{o}}}}<1.
\]
As a result, 
\begin{equation}
V_{\text{o}}\approx\frac{k_{2}}{\epsilon\left\Vert \boldsymbol{\tilde{\xi}}{_{\text{o}}}\right\Vert }>\frac{k_{2}}{\epsilon\left(\gamma r_{\text{s}}+{{r}_{\text{o}}}\right)}\label{Property}
\end{equation}
which is very large as $\epsilon>0$ is chosen very small. 
\end{itemize}
The objective of the designed velocity command is to make $V_{\text{w}}\left(\boldsymbol{\tilde{\xi}}{_{\text{wp}}}\right)$
and$\ V_{\text{o}}\left(\left\Vert \boldsymbol{\tilde{\xi}}{_{\text{o}}}\right\Vert \right)$
be zero or as small as possible. According to \textit{Lemma 1}\textbf{
}and Property (ii), this implies $\left\Vert \boldsymbol{\tilde{\xi}}{_{\text{wp}}}\right\Vert \rightarrow0$
and$\ \left\Vert \boldsymbol{\tilde{\xi}}{_{\text{o}}}\right\Vert \geq r_{\text{a}}$.
Namely, the multicopter will arrive at the goal waypoint ${{\mathbf{p}}_{\text{wp}}}$
and does not collide with the obstacle.

\subsection{Controller Design}

The velocity command for the multicopter modeled in (\ref{positionmodel_ab_con_i})
is designed as 
\begin{equation}
\mathbf{v}_{\text{c}}=\mathbf{-}\text{sat}\left(\text{sat}\left(k_{1}\boldsymbol{\tilde{\xi}}_{\text{wp}},v_{\text{m}}\right)-a_{\text{o}}\boldsymbol{\tilde{\xi}}_{\text{o}},v_{\text{m}}\right)\label{control_p2_1}
\end{equation}
where $k_{1}>0$ and\footnote{$a_{\text{o}}\geq0$ according to the property (i) of $V_{_{\text{o}}}.$}
\begin{equation}
a_{\text{o}}=-\frac{\partial V_{\text{o}}}{\partial\left\Vert \boldsymbol{\tilde{\xi}}_{\text{o}}\right\Vert }\frac{1}{\left\Vert \boldsymbol{\tilde{\xi}}_{\text{o}}\right\Vert }.\label{aij}
\end{equation}

\textbf{Remark 3}. The saturation constraint term sa${\text{t}}\left(k_{1}\boldsymbol{\tilde{\xi}}{_{\text{wp}}},v_{\text{m}}\right)$
in (\ref{control_p2_1})\ is very necessary. Without the saturation,
the velocity command (\ref{control_p2_1})\ becomes 
\[
\mathbf{v}_{\text{c}}=\mathbf{-}\text{sa}{\text{t}}\left(k_{1}\boldsymbol{\tilde{\xi}}{_{\text{wp}}}-a{_{\text{o}}}\boldsymbol{\tilde{\xi}}{_{\text{o}}},{v_{\text{m}}}\right).
\]
In this case, if the initial value $\boldsymbol{\tilde{\xi}}{_{\text{wp}}}\left(0\right)$
is very large, then the term $k_{1}\boldsymbol{\tilde{\xi}}{_{\text{wp}}}$
will dominate until the multicopter is very close to the obstacle
so that $a{_{\text{o}}}\boldsymbol{\tilde{\xi}}{_{\text{o}}}$ can
dominate. In this case, $a_{\text{o}}\approx\frac{k_{2}}{\epsilon}\frac{1}{\left\Vert \boldsymbol{\tilde{\xi}}{_{\text{o}}}\right\Vert ^{3}},$
which will increase sharply when the multicopter is very close to
the obstacle. At that time, the multicopter will start to change the
velocity to avoid a conflict. In practice, it may be too late by taking
various uncertainties into consideration. The use of the maximum speeds
$v_{\text{m}}$ in the term sa${\text{t}}\left(k_{1}\boldsymbol{\tilde{\xi}}{_{\text{wp}}},v_{\text{m}}\right)$
of the velocity command (\ref{control_p2_1}) will avoid such a danger.

\begin{figure}[h]
\begin{centering}
\includegraphics{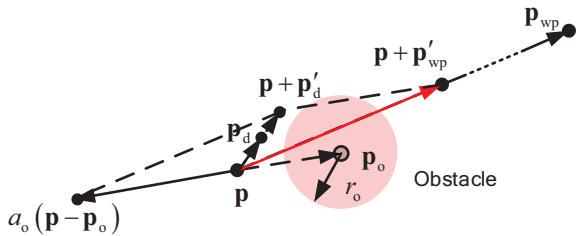} 
\par\end{centering}
\caption{Desired position generated for obstacle avoidance.}
\label{pdobstacleavoidance}
\end{figure}

\textbf{Remark 4}. The controller (\ref{control_p2_1}) can be written
as a proportional (P) controller as 
\[
\mathbf{v}_{\text{c}}={{\mathbf{p}}_{\text{d}}}-{\mathbf{p}}
\]
where 
\begin{equation}
{{\mathbf{p}}_{\text{d}}}={\mathbf{p}}-\text{sa}{\text{t}}\left(\text{sa}{\text{t}}\left(k_{1}\boldsymbol{\tilde{\xi}}{_{\text{wp}}},v_{\text{m}}\right)-a{_{\text{o}}}\boldsymbol{\tilde{\xi}}{_{\text{o}}},{v_{\text{m}}}\right).\label{pd}
\end{equation}
In order to figure out the physical meaning of ${{\mathbf{p}}_{\text{d}}}$,
suppose that $\mathbf{v}=\mathbf{0,}$ $\mathbf{v}_{\text{o}}=\mathbf{0}$
and ${\mathbf{p}}$ is close to the obstacle, namely $\left\Vert {\mathbf{p}}-{\mathbf{p}}_{\text{o}}\right\Vert <r_{\text{a}}.$\textit{
}This implies $a_{\text{o}}{\mathbf{\tilde{p}}}_{\text{o}}\neq\mathbf{0}\ $according
to the property of $V_{\text{o}}$. Then (\ref{pd}) becomes 
\begin{equation}
{{\mathbf{p}}_{\text{d}}}=\mathbf{p}-\text{sa}{\text{t}}\left(\text{sa}{\text{t}}\left(k_{1}\mathbf{\tilde{p}}{_{\text{wp}}},v_{\text{m}}\right)-a_{\text{o}}{\mathbf{\tilde{p}}}_{\text{o}},{v_{\text{m}}}\right).\label{pd2}
\end{equation}
Furthermore, let 
\[
k_{1}=k_{2}=1.
\]
Then ${{\mathbf{p}}_{\text{d}}}$ has a concise form 
\[
{{\mathbf{p}}_{\text{d}}}=\mathbf{p}+\text{sa}{\text{t}}\left({\mathbf{p}}_{\text{d}}^{\prime},{v_{\text{m}}}\right)
\]
where ${{\mathbf{p}}_{\text{d}}^{\prime}}={{\mathbf{p}}_{\text{wp}}^{\prime}}+a_{\text{o}}\left(\mathbf{p}-\mathbf{p}_{\text{o}}\right)\ $and
${{\mathbf{p}}_{\text{wp}}^{\prime}}=$sat$\left(\mathbf{p}{_{\text{wp}}}-{\mathbf{p}},v_{\text{m}}\right).$
The physical meaning of ${{\mathbf{p}}_{\text{d}}}$ and ${{\mathbf{p}}_{\text{d}}^{\prime}}$
is shown in Figure \ref{pdobstacleavoidance}. As shown, the contribution
$\mathbf{p}{_{\text{wp}}}-{\mathbf{p}}$ is saturated by the term
sa{{t}}$\left(\mathbf{p}{_{\text{wp}}}-{\mathbf{p}},v_{\text{m}}\right).$
Otherwise, ${{\mathbf{p}}_{\text{d}}}$ will still point to the obstacle
until the multicopter is very close to the obstacle. 

\textbf{Remark 5}. Although the property of the obstacle avoidance
is proved, the case such as $\left\Vert \boldsymbol{\tilde{\xi}}{_{\text{o}}}\right\Vert <r_{\text{s}}+{{r}_{\text{o}}}$
may still happen in practice due to unpredictable uncertainties. For
example, the obstacle cannot be detected until it is very close to
the multicopter. However, this may not imply that the multicopter
has a collision with the obstacle physically. In this case, 
\[
a_{\text{o}}\approx\frac{k_{2}}{\epsilon}\frac{1}{\left\Vert \boldsymbol{\tilde{\xi}}{_{\text{o}}}\right\Vert ^{3}}.
\]
Since $\epsilon$ is chosen to be sufficiently small, the term $a_{\text{o}}\boldsymbol{\tilde{\xi}}{_{\text{o}}}$
will dominate so that the velocity command $\mathbf{v}_{\text{c}}$
becomes 
\[
\mathbf{v}_{\text{c}}\approx\text{sa}{\text{t}}\left(\frac{k_{2}}{\epsilon}\frac{1}{\left\Vert \boldsymbol{\tilde{\xi}}{_{\text{o}}}\right\Vert ^{2}}\frac{\boldsymbol{\tilde{\xi}}{_{\text{o}}}}{\left\Vert \boldsymbol{\tilde{\xi}}{_{\text{o}}}\right\Vert },{v_{\text{m}}}\right).
\]
This implies that, by recalling (\ref{obmodel}), $\left\Vert \boldsymbol{\tilde{\xi}}{_{\text{o}}}\right\Vert $
will be increased fast with its maximum speed. This implies that the
multicopter will keep away from the obstacle immediately.

\subsection{Stability Analysis}

In order to investigate the convergence to the goal waypoint and the
obstacle avoidance behaviour, a function is defined as follows 
\begin{equation}
{{V}_{1}}=V_{\text{w}}+V_{\text{o}}\label{V1}
\end{equation}
where $V_{\text{w}}\ $is defined in (\ref{Vw}), and $V_{\text{o}}\ $is
defined in (\ref{Vo}). Before introducing the main result, three
lemmas are needed.

\textbf{Lemma 2}. Under \textit{Assumption 2},\textbf{ }for (\ref{positionmodel_ab_con_i}),
if the velocity input $\mathbf{v}_{\text{c}}$ is designed as in (\ref{control_p2_1})
and ${v_{\text{m}}}>v_{\text{o}}$, then there exist sufficiently
small $\epsilon,\epsilon_{\text{s}}>0$ and any $\gamma>1$ in $a_{0}$
such that$\ \left\Vert \boldsymbol{\tilde{\xi}}{_{\text{o}}}\left(t\right)\right\Vert >r_{\text{s}}+{{r}_{\text{o}},}$
$t\in\left[0,\infty\right).$

\textit{Proof}. See \emph{Appendix}. $\square$ 
\begin{figure}[h]
\begin{centering}
\includegraphics[scale=0.8]{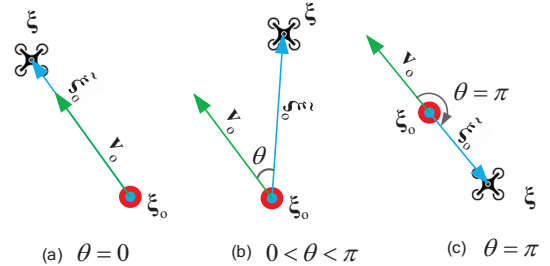} \vspace{-2em}
 
\par\end{centering}
\caption{Relationship between filtered position error and the velocity of the
obstacle.}
\label{Relobvsuav}
\end{figure}

As shown in Figure \ref{Relobvsuav}, let 
\begin{equation}
\cos\theta=\frac{\boldsymbol{\tilde{\xi}}_{\text{o}}^{\text{T}}\mathbf{v}{_{\text{o}}}}{\left\Vert \mathbf{v}{_{\text{o}}}\right\Vert \left\Vert \boldsymbol{\tilde{\xi}}{_{\text{o}}}\right\Vert }\label{costheta}
\end{equation}
where $\theta\in\left[0,\pi\right]\ $is the angle between $\boldsymbol{\tilde{\xi}}_{\text{o}}$
and $\mathbf{v}{_{\text{o}}}${.} The positive direction of the
rotation is clockwise without loss of generality. The next lemma is
to show that the relationship between the obstacle and the multicopter
will tend to the case shown in Figure \ref{Relobvsuav}(c).

\textbf{Lemma 3}. Under \textit{Assumptions 1-3},\textbf{ }for (\ref{positionmodel_ab_con_i}),
if the velocity input $\mathbf{v}{_{\text{c}}}$ is designed as in
(\ref{control_p2_1}) and $\mathbf{v}{_{\text{o}}}\neq\mathbf{0}$,
then $\underset{t\rightarrow\infty}{\lim}\theta\left(t\right)=\pi{\ }${for}
{almost} all $\boldsymbol{\tilde{\xi}}_{\text{o}}\left(0\right).$

\textit{Proof}. See \emph{Appendix}. $\square$

\textbf{Lemma 4}. If $0<\rho\mathbf{I}_{n}<\Lambda=\Lambda^{\text{T}}\in
\mathbb{R}
^{n\times n}\ $is$\ $a symmetric positive-definite matrix, and $\mathbf{X}\in
\mathbb{R}
^{n\times n}\mathbf{\ }$is a symmetric matrix with rank$\left(\mathbf{X}\right)\leq n-1,$
then $\lambda_{\max}\left(\Lambda+\mathbf{X}\right)>\rho>0,$ where
$\lambda_{\max}\left(\Lambda+\mathbf{X}\right)$ denotes the maximum
eigenvalue of $\Lambda+\mathbf{X}.$

\textit{Proof.} See \emph{Appendix}. $\square$

With \emph{Lemmas 1-4} in hand, we can state the main result.

\textbf{Theorem 1}. Under \textit{Assumptions 1-3},\textit{ }suppose
that the velocity input for the multicopter is designed as in (\ref{control_p2_1}).
Then there exist sufficiently small $\epsilon,\epsilon_{\text{s}}>0$
and $\gamma>1$ in $a_{0}$ such that $\lim_{t\rightarrow\infty}\left\Vert {{\mathbf{\tilde{p}}}_{\text{wp}}}\left(t\right)\right\Vert =0$
and $\left\Vert \boldsymbol{\tilde{\xi}}{_{\text{o}}}\left(t\right)\right\Vert >r_{\text{s}}+{{r}_{\text{o}}},$
$t\in\lbrack0,\infty)$ for almost\footnote{For all initial conditions $\boldsymbol{\tilde{\xi}}{_{\text{wp}}(0),}$
$\mathbf{p}_{\text{wp,}i}$ is a stable equilibrium with probability
1, and other equilibriums are unstable with probability 1.} all $\boldsymbol{\tilde{\xi}}{_{\text{wp}}(0)}$ when $v_{\text{m}}>v{_{\text{o}}}${.}

\textit{Proof}. By\textit{ Lemma 2}, there exist sufficiently small
$\epsilon,\epsilon_{\text{s}}>0$ and any $\gamma>1$ in $a_{0}$
such that$\ \left\Vert \boldsymbol{\tilde{\xi}}{_{\text{o}}}\left(t\right)\right\Vert >r_{\text{s}}+{{r}_{\text{o}},}$
$t\in\left[0,\infty\right){,}$ namely the obstacle avoidance can
be achieved. In order to investigate the convergence to the goal waypoint,
a function is defined as in (\ref{V1}). According to the line integrals
of vector fields\cite[p. 911]{Thomas(2009)}, one has 
\[
{{V}_{1}}=
{\displaystyle \int\nolimits _{0}^{t}}
\text{sa}{\text{t}}\left(k_{1}\boldsymbol{\tilde{\xi}}{_{\text{wp}}},v_{\text{m}}\right)^{\text{T}}\boldsymbol{\dot{\tilde{\xi}}}{_{\text{wp}}}\text{d}\tau+V_{\text{o}}.
\]
The derivative of ${{V}_{1}}$ along the solution to (\ref{wpmodel})
and (\ref{obmodel}) is 
\[
{{\dot{V}}_{1}}=\left(\text{sa}{\text{t}}\left(k_{1}\boldsymbol{\tilde{\xi}}{_{\text{wp}}},v_{\text{m}}\right)-a_{\text{o}}\boldsymbol{\tilde{\xi}}{_{\text{o}}}\right)^{\text{T}}\mathbf{v}_{\text{c}}+a_{\text{o}}\boldsymbol{\tilde{\xi}}_{\text{o}}^{\text{T}}\mathbf{v}_{\text{o}}
\]
where $a_{\text{o}}$ are defined in (\ref{aij}){ and }$\mathbf{a}_{\text{o}}=\mathbf{v}_{\text{o}}$
according to \textit{Assumption 1}. By using the velocity input (\ref{control_p2_1}),
${{\dot{V}}_{1}}$ becomes 
\begin{align}
{{\dot{V}}_{1}} & =-\left(\text{sa}{\text{t}}\left(k_{1}\boldsymbol{\tilde{\xi}}{_{\text{wp}}},v_{\text{m}}\right)-a_{\text{o}}\boldsymbol{\tilde{\xi}}{_{\text{o}}}\right)^{\text{T}}\nonumber \\
 & \text{ \ \ }\cdot\text{sa}{\text{t}}\left(\text{sa}{\text{t}}\left(k_{1}\boldsymbol{\tilde{\xi}}{_{\text{wp}}},v_{\text{m}}\right)-a_{\text{o}}\boldsymbol{\tilde{\xi}}{_{\text{o}}},{v_{\text{m}}}\right)+a_{\text{o}}\boldsymbol{\tilde{\xi}}_{\text{o}}^{\text{T}}\mathbf{v}_{\text{o}}.\label{dV2}
\end{align}

In the following, we will discuss the stability of two cases, namely
$\mathbf{v}{_{\text{o}}}=\mathbf{0}$ and $\mathbf{v}{_{\text{o}}}\neq\mathbf{0.}$
\begin{itemize}
\item \textbf{Property of function} $V_{1}.$ Before applying \textit{invariant
set theorem }\cite{Slotine(1991)}, we will study the property of
function $V_{1}$. Let $\Omega=\left\{ \left.\boldsymbol{\xi}\right\vert {V_{1}}\left(\boldsymbol{\xi}\right)\leq l_{0}\right\} ,$
$l_{0}>0.$ According to \textit{Lemma 2}, $V_{\text{o}}>0.$ Therefore,
${V_{1}}\left(\boldsymbol{\xi}\right)\leq l_{0}$ implies $V_{\text{w}}\leq l_{0}.$
Furthermore, according to \textit{Lemma 1(iii)}, $\Omega$ is bounded.
When $\left\Vert \boldsymbol{\xi}\right\Vert \rightarrow\infty,$
then $V_{\text{w}}\rightarrow\infty$ according to \textit{Lemma 1(ii)},
namely ${V}_{1}\rightarrow\infty.$
\item \textbf{Case 1: Obstacle is static}. (i) If $\mathbf{v}{_{\text{o}}}=\mathbf{0}$,
then (\ref{dV2}) becomes 
\begin{align}
{{\dot{V}}_{1}}= & -\left(\text{sa}{\text{t}}\left(k_{1}\boldsymbol{\tilde{\xi}}{_{\text{wp}}},v_{\text{m}}\right)-a_{\text{o}}\boldsymbol{\tilde{\xi}}{_{\text{o}}}\right)^{\text{T}}\nonumber \\
 & \cdot\text{sa}{\text{t}}\left(\text{sa}{\text{t}}\left(k_{1}\boldsymbol{\tilde{\xi}}{_{\text{wp}}},v_{\text{m}}\right)-a_{\text{o}}\boldsymbol{\tilde{\xi}}{_{\text{o}}},{v_{\text{m}}}\right).\label{dV21}
\end{align}
In the following, we will show the convergence to the goal waypoint.
The invariant set theorem is used to do the following analysis\textit{.
}Now, ${{\dot{V}}_{1}}={0}$ if and only if 
\begin{equation}
\text{sa}{\text{t}}\left(k_{1}\boldsymbol{\tilde{\xi}}{_{\text{wp}}},v_{\text{m}}\right)-a_{\text{o}}\boldsymbol{\tilde{\xi}}{_{\text{o}}}=\mathbf{0}\label{equilibrium0_v}
\end{equation}
according to (\ref{dV21}). Then $\mathbf{v}_{\text{c}}=\mathbf{0\ }$according
to (\ref{control_p2_1}). Consequently, by (\ref{positionmodel_ab_con_i}),
the system cannot get ``stuck''\ at an equilibrium value except
for $\mathbf{v}=\mathbf{0}$. The equation (\ref{equilibrium0_v})
can be further written as 
\begin{equation}
k_{1}\mathbf{\tilde{p}}{_{\text{wp}}}-a_{\text{o}}{\mathbf{\tilde{p}}{_{\text{o}}}}=\mathbf{0.}\label{ch13-equilibrium0}
\end{equation}
According to equation (\ref{ch13-equilibrium0}), the equilibrium
point $\mathbf{p}^{\ast}$ obviously sits on the straight line through
${{\mathbf{p}}_{\text{wp}}}$ and ${{\mathbf{p}}_{\text{o}}}$. As
shown in Figure \ref{obstacleavoidance}(a), the straight line is
divided into: ``A half-line'', ``B segment''\ and ``C half-line''.
Obviously, the equilibrium point $\mathbf{p}^{\ast}$ cannot be on
``A half-line''\ and ``B segment''\ except for $\mathbf{p}^{\ast}=\mathbf{p}{_{\text{wp}}}$.
Without loss of generality, it is assumed that the solution lying
on ``C half-line''\ is $\mathbf{p}^{\ast}=\mathbf{\bar{p}}$, as
shown in Figure \ref{obstacleavoidance}(b). 
\begin{figure}[h]
\begin{centering}
\includegraphics[scale=0.8]{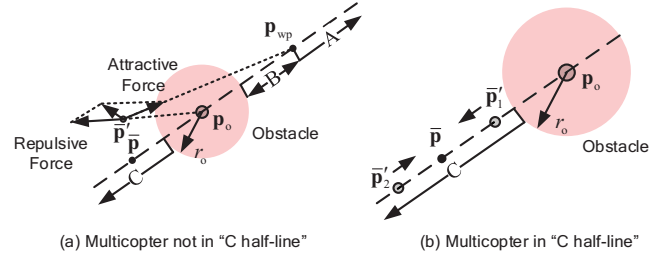} \vspace{-2em}
 
\par\end{centering}
\caption{Desired position generated for obstacle avoidance.}
\label{obstacleavoidance}
\end{figure}

\ In the following, we will analyze the stability of $\mathbf{\bar{p}}$
on ``C half-line''. By (\ref{ch13-equilibrium0}), we have 
\begin{equation}
\frac{a_{\text{o}}}{k_{1}}=\frac{\left\Vert \mathbf{\tilde{p}}{_{\text{wp}}}\right\Vert }{\left\Vert {\mathbf{\tilde{p}}{_{\text{o}}}}\right\Vert }\label{a0k1}
\end{equation}
at the equilibrium point $\mathbf{p}=\mathbf{\bar{p},v=0.}$ Since
the solution lying on ``C half-line'', $\left\Vert \mathbf{\tilde{p}}{_{\text{wp}}}\right\Vert =\left\Vert {\mathbf{\tilde{p}}{_{\text{o}}}}\right\Vert +\left\Vert \mathbf{p}{_{\text{wp}}-}\mathbf{p}{_{\text{o}}}\right\Vert .$
Therefore, by (\ref{a0k1}), we have 
\begin{equation}
a_{\text{o}}>k_{1}{.}\label{inequ}
\end{equation}
Let 
\[
\mathbf{f}\left(\boldsymbol{\xi}\right)=-k_{1}\boldsymbol{\tilde{\xi}}{_{\text{wp}}}+a_{\text{o}}\boldsymbol{\tilde{\xi}}{_{\text{o}}.}
\]
Then 
\[
\left.\frac{\partial\mathbf{f}\left(\boldsymbol{\xi}\right)}{\partial\boldsymbol{\xi}}\right\vert _{\mathbf{p}=\mathbf{\bar{p},v=0}}=\left.\left(\Lambda+\mathbf{B}\right)\right\vert _{\mathbf{p}=\mathbf{\bar{p},v=0}}
\]
where 
\[
\Lambda=\left(-k_{1}+a_{\text{o}}\right)\mathbf{I}_{2},\text{ }\mathbf{B}=\left.\frac{\partial a_{\text{o}}}{\partial\left\Vert \boldsymbol{\tilde{\xi}}{_{\text{o}}}\right\Vert }\boldsymbol{\tilde{\xi}}{{_{\text{o}}}\boldsymbol{\tilde{\xi}}_{\text{o}}^{\text{T}}}\right\vert _{\mathbf{p}=\mathbf{\bar{p},v=0}}{.}
\]
Since rank$\left(\mathbf{B}\right)=1<2\ $and\ $0<\Lambda=\Lambda^{\text{T}}\in
\mathbb{R}
^{n\times n}\ $is$\ $a symmetric positive-definite matrix by the fact (\ref{inequ}),
we have $\lambda_{\max}\left(\left.\left(\Lambda+\mathbf{B}\right)\right\vert _{\mathbf{p}=\mathbf{\bar{p},v=0}}\right)>0\ $according
to \emph{Lemma 4}. This implies, at the equilibrium point $\mathbf{p}=\mathbf{\bar{p},}$
$\mathbf{v=0}$, the dynamics 
\[
\boldsymbol{\dot{\xi}}=\mathbf{f}\left(\boldsymbol{\xi}\right)
\]
is unstable because one of eigenvalues is positive. This does not
imply the other eigenvalue is positive as well. If the other eigenvalue
is negative, then the multicopter only can be stable in one dimensional
space, which the measure is 0 on a 2D space. Therefore, we can conclude
this proof when $\mathbf{v}{_{\text{o}}}=\mathbf{0}$.
\item \textbf{Case 2: Obstacle is moving}. The proof is quite different
from the proof above because, if $\mathbf{v}{_{\text{o}}}\neq\mathbf{0}$,
the equilibrium point as well as its property are not easy to get
from\ (\ref{dV2}) directly. If $\mathbf{v}{_{\text{o}}}\neq\mathbf{0}$,
then\textit{ }$\underset{t\rightarrow\infty}{\lim}\theta\left(t\right)=\pi{\ }${for}
{almost} all $\boldsymbol{\tilde{\xi}}_{\text{o}}\left(0\right)$
by \textit{Lemma 3}. This implies that there exists a time $t_{1}>0$
such that $\left\Vert \mathbf{v}{_{\text{o}}}\right\Vert \left\Vert \boldsymbol{\tilde{\xi}}{_{\text{o}}}\right\Vert \cos\theta=\boldsymbol{\tilde{\xi}}_{\text{o}}^{\text{T}}\left({t}\right)\mathbf{v}{_{\text{o}}}\leq{0\ }${for
}${t\geq t}_{1}.$ When $t_{1}>0$, (\ref{dV2}) becomes 
\begin{align}
{{\dot{V}}_{1}}\leq & -\left(\text{sa}{\text{t}}\left(k_{1}\boldsymbol{\tilde{\xi}}{_{\text{wp}}},v_{\text{m}}\right)-a_{\text{o}}\boldsymbol{\tilde{\xi}}{_{\text{o}}}\right)^{\text{T}}\nonumber \\
 & \cdot\text{sa}{\text{t}}\left(\text{sa}{\text{t}}\left(k_{1}\boldsymbol{\tilde{\xi}}{_{\text{wp}}},v_{\text{m}}\right)-a_{\text{o}}\boldsymbol{\tilde{\xi}}{_{\text{o}}},{v_{\text{m}}}\right)\label{dV3}
\end{align}
\end{itemize}
where the key step is to eliminate the term $a_{\text{o}}\boldsymbol{\tilde{\xi}}_{\text{o}}^{\text{T}}\mathbf{v}_{\text{o}}$
of (\ref{dV2}) by \textit{Lemma 3}. The following proof is similar
to that of \textit{Case 1. }$\square$

\textbf{Remark 5}. We further explain why the multicopter cannot lie
on ``C half-line''\ intuitively. As shown in Figure \ref{obstacleavoidance}(a),
if the multicopter deviates from ``C half-line'', for instance,
it reaches $\mathbf{p}^{\ast}=\mathbf{{\bar{p}}^{\prime}}$, then
the sum of the attractive force and the repulsive force will make
the multicopter further keep away from the ``C half-line''. On the
other hand, as shown in Figure \ref{obstacleavoidance}(b), when the
multicopter gets close to the obstacle along ``C half-line'', for
instance, it reaches $\mathbf{p}^{\ast}=\mathbf{\bar{p}}_{1}^{\prime}$,
the repulsive force will dominate because of the term $a_{\text{o}}\approx\frac{k_{2}}{\epsilon}\left/\left\Vert \boldsymbol{\tilde{\xi}}{_{\text{o}}}\right\Vert ^{3}\right.$
in the equation (\ref{control_p2_1}). As a result, a relatively large
repulsive force will push it back to $\mathbf{\bar{p}}$. By contrary,
if the multicopter reaches $\mathbf{p}^{\ast}={\mathbf{{\bar{p}}}}_{2}^{\prime}$,
the attractive force will dominate. As a result, a relatively large
attractive force will pull it back towards $\mathbf{\bar{p}}$. However,
in practice, a multicopter will never strictly stay on ``C half-line'',
namely the measure of ``C half-line''\ in 2D space equals 0 or
the stability probability is 0. Therefore, any small deviation from
the ``C half-line''\ will drive the multicopter away from $\mathbf{\bar{p}}$.
In conclusion, the solution $\mathbf{p}^{\ast}={{\mathbf{p}}_{\text{wp}}}$
lying on ``A half-line''\ is the only stable equilibrium point.
It is also globally asymptotically stable with probability 1, namely
$\lim_{t\rightarrow\infty}\left\Vert {{\mathbf{\tilde{p}}}_{\text{wp}}}\left(t\right)\right\Vert =0$
for almost ${{\mathbf{\tilde{p}}}_{\text{wp}}(0)}$.

\textbf{Remark 6}. It is necessary to assume $v_{\text{m}}>v{_{\text{o}}}$.
Otherwise, in the worst case, the collision cannot be avoided. For
example, we can choose the obstacle dynamic as 
\begin{equation}
\boldsymbol{\dot{\xi}}{_{\text{o}}}=\boldsymbol{\dot{\xi}}-{\epsilon}\frac{\boldsymbol{\xi}{_{\text{o}}}-\boldsymbol{\xi}}{\left\Vert \boldsymbol{\xi}{_{\text{o}}}-\boldsymbol{\xi}\right\Vert }\label{Remark4}
\end{equation}
where $\left\Vert \boldsymbol{\dot{\xi}}{_{\text{o}}}\right\Vert \leq\left\Vert \boldsymbol{\dot{\xi}}\right\Vert +{\epsilon}_{1}\leq{v_{\text{m}}+\epsilon}_{1}$
with ${\epsilon}_{1}>{0.}$ From (\ref{Remark4}), it is easy to see
$\left\Vert \boldsymbol{\xi}_{\text{o}}\left(t\right)-\boldsymbol{\xi}\left(t\right)\right\Vert <r_{\text{s}}+r_{\text{o}}$
within a finite time no matter how small ${\epsilon}_{1}$ is. This
implies that the obstacle is chasing after the multicopter and then
hits it finally. It is worth pointing out the non-cooperate obstacle
model considered in this paper has an unknown and unpredictable trajectory,
but it has no intention of intercepting the multicopter (such as the
obstacle always move to the line between the aircraft and the target
point). The strategies of how to avoid such obstacles are further
work. In theory, if the proposed obstacle model in this paper is used
for analysis (with the exception of maximum speed, there are no more
restrictions for them, such as the turning radius), it would be impossible
to avoid these intercepted obstacles if the condition $v_{\text{m}}>v{_{\text{o}}}$
do not hold.

\subsection{Result Extension to Multiple Moving Obstacles}

\begin{figure}[h]
\begin{centering}
\includegraphics[scale=0.5]{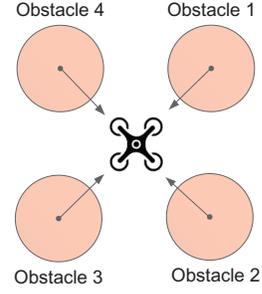} 
\par\end{centering}
\caption{A multicopter surrounded by four obstacles.}
\label{surround}
\end{figure}

The avoidance case with multiple non-cooperative moving obstacles
is complex. Under some initial conditions, a multicopter cannot avoid
collision with obstacles no matter what a controller uses, such as
a case shown in Figure \ref{surround}. However, under some cases,
the multiple moving\ obstacle avoidance control problem can be solved
based on the results on the one moving obstacle avoidance control
problem directly.

\textbf{(1) Multiple Parallel Moving Obstacles with the Same Velocity}

\begin{figure}[h]
\begin{centering}
\includegraphics[scale=0.7]{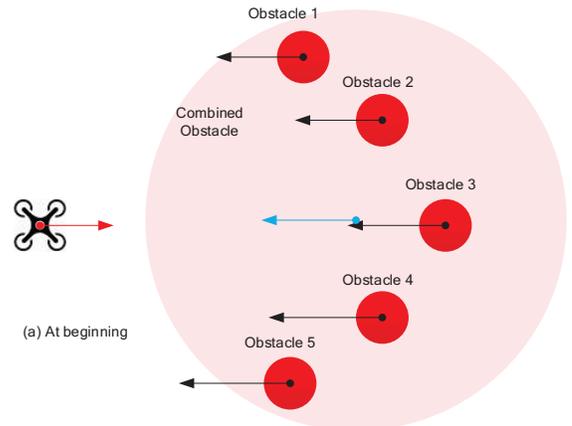} 
\par\end{centering}
\caption{Multiple clustered obstacles taken as a combined obstacle.}
\label{combinedobstacle}
\end{figure}

In practice, a multicopter will face a group of wild gooses, which
move with a constant velocity are parallel with each other. In order
to adopt the proposed method, the clustered wild goose can be taken
as one combined obstacle (as shown in Figure \ref{combinedobstacle})
with appropriate center position $\mathbf{p}_{\text{o,c}}$, velocity
$\mathbf{v}_{\text{o,c}}$ and radius $r_{\text{o,c}}$. Define 
\begin{align*}
\boldsymbol{\xi}_{\text{o,c}} & =\mathbf{p}_{\text{o,c}}+\frac{1}{{l}}\mathbf{v}_{\text{o,c}}\\
\boldsymbol{\tilde{\xi}}_{\text{o,c}} & =\boldsymbol{\xi}-\boldsymbol{\xi}_{\text{o,c}}.
\end{align*}
The combined obstacle satisfies $\max\left\Vert \boldsymbol{\dot{\xi}}_{\text{o,c}}\right\Vert \leq v_{\text{o}}$,
$k=1,\cdots,{N}$. Similarly, we have \textit{Assumption 2'} to replace
with \textit{Assumption 2 }in the following.

\textbf{Assumption 2'}. The multicopter's initial filtered position
$\boldsymbol{\tilde{\xi}}{_{\text{o}}}\left(0\right)\in{{\mathbb{R}}^{2}}$
satisfies 
\[
\left\Vert \boldsymbol{\tilde{\xi}}_{\text{o,c}}\left(0\right)\right\Vert >r_{\text{s}}+r_{\text{o,c}}
\]
and $\left\Vert \mathbf{v}\left(0\right)\right\Vert \leq v_{\text{m}}.$

Based on \textit{Assumptions 1,2',3}, the multiple moving obstacle
problem can be degraded to the one moving obstacle avoidance control
problem. As a result, the proposed method can still work. 

\textbf{(2) Multiple Non-Parallel Moving Obstacles}

At the same altitude, there are ${N}$ moving obstacles 
\[
\mathcal{O}_{\text{o,}k}=\left\{ \mathbf{x}\in{{\mathbb{R}}^{2}}\left\vert \left\Vert \mathbf{x}-{{\mathbf{p}}_{\text{o,}k}}\right\Vert \leq{{r}_{\text{o,}k}}\right.\right\} 
\]
where ${{\mathbf{p}}_{\text{o,}k}}\in{{\mathbb{R}}^{2}}$ is the center
position of the $k$th obstacle, $\mathbf{v}{_{\text{o,}k}={\mathbf{\dot{p}}}_{\text{o,}k}}\in{{\mathbb{R}}^{2}}$
is the velocity of the $k$th obstacle, $r_{\text{o,}k}>0$ is the
radius of the $k$th obstacle, $k=1,\cdots,{N}$. Define 
\begin{align*}
\boldsymbol{\xi}_{\text{o,}k} & =\mathbf{p}_{\text{o,}k}+\frac{1}{{l}}\mathbf{v}_{\text{o,}k}\\
\boldsymbol{\tilde{\xi}}_{\text{o,}k} & =\boldsymbol{\xi}-\boldsymbol{\xi}_{\text{o,}k}.
\end{align*}
These obstacles satisfy $\max\left\Vert \boldsymbol{\dot{\xi}}_{\text{o,}k}\right\Vert \leq v_{\text{o}}$,
$k=1,\cdots,{N}$. To extend the conclusions\textit{ }to\textit{\ }multiple
moving obstacles, we have \textit{Assumption 2''} to replace with
\textit{Assumption 2 }in the following.

\textbf{Assumption 2''}. The multicopter's initial filtered position
$\boldsymbol{\tilde{\xi}}{_{\text{o}}}\left(0\right)\in{{\mathbb{R}}^{2}}$
satisfies 
\[
\left\Vert \boldsymbol{\tilde{\xi}}_{\text{o,}k}\left(0\right)\right\Vert >r_{\text{s}}+{{r}_{\text{o,}k}}
\]
and $\left\Vert \mathbf{v}\left(0\right)\right\Vert \leq v_{\text{m}},$
$k=1,\cdots,{N}$. The distances among obstacles satisfy 
\begin{equation}
\left\Vert {\mathbf{p}}_{\text{o,}i}\left(t\right)-{\mathbf{p}}_{\text{o,}j}\left(t\right)\right\Vert \geq2{r}_{\text{a}}+\left({{r}_{\text{o,}i}}+{{r}_{\text{o,}j}}\right),i\neq j\label{distance}
\end{equation}
then the multicopter cannot be close to two obstacles at the same
time, namely
\[
\mathcal{A}\cap\mathcal{O}_{\text{o,}i}\neq\varnothing,\mathcal{A}\cap\mathcal{O}_{\text{o,}j}\neq\varnothing
\]
where $i\neq j,$ $i,j=1,\cdots,N.$

Based on \textit{Assumptions 1,2'',3}, the multicopter will encounter
only one obstacle at any time so that the proposed method can still
work. Here, each obstacle moves with a constant velocity, not paralleling
with each other. After enough time, non-parallel obstacles must be
separated far enough with each other. In this case, only one obstacle
at most needs to be considered for the multicopter after enough time
because the others are out of its avoidance area. The multiple moving
obstacles problem can be degraded to the one moving obstacle avoidance
control problem. As a result, the proposed method can still work.
Let us discuss on the condition (\ref{distance}). If the condition
(\ref{distance}) is not satisfied at some time, the multicopter is
also safe if it is not at the place where two obstacls do not satisfy
(\ref{distance}). Even if the multicopter will be the place, it can
take the these dangerous obstacles as one combined obstacle similar
to the case of the multiple parallel moving obstacles. The strategies
how to combine multiple obstacles are the further work. 
\begin{figure}[h]
\begin{centering}
\includegraphics[scale=0.7]{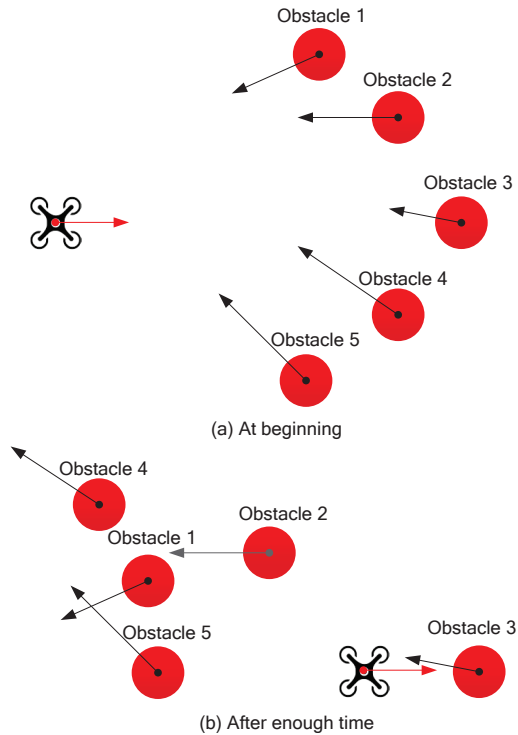} 
\par\end{centering}
\caption{Multiple\textbf{ }non-parallel obstacles at the beginning and after
enough time.}
\label{MultipleNonCooperativeObstacles}
\end{figure}

\textbf{Remark 7}. If the multicopter and obstacles are \emph{cooperative},
namley they make avoidance with each other, then multiple moving obstacles
can keep away with each other and go to their waypints finally. The
dilemma shown in Figure \ref{surround} and the velocity constraint
will not exist for the cooperative case. The formal proof is omitted
here because of limited space. A simple explaination on the dilemma
is that the other\ cooperative obstacles can keep away so that they
satisfy \textit{Assumption 2''.} On ther other hand, the explaination
on the velocity constraint is that the fast one is taken as the multicopter
while the other taken as obstacles.

\section{Simulation and Experiment}

Simulations and experiments are given in the following to show the
effectiveness of the proposed method, where a video about simulations
and experiments is available on https://youtu.be/0kedvrRXUd8 or https://suo.im/6wObz7.

\subsection{Simulation}

We design three different scenarios to show the performance of the
proposed controller. In the first scenario, the effectiveness of the
proposed method to avoid one obstacle is shown. In the last two scenarios,
the feasibility of the result extension to multiple obstacles is demonstrated.

\textbf{(1) Simulation with One-on-One Non-Cooperative Obstacle}

\begin{figure}[ptb]
\begin{centering}
\includegraphics[scale=0.6]{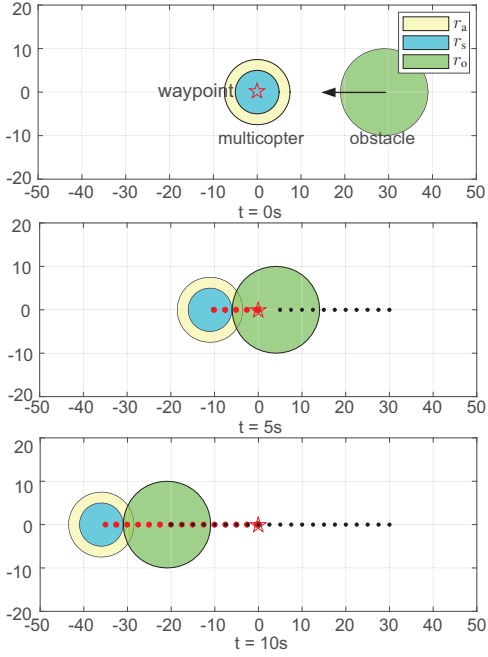}
\par\end{centering}
\caption{Positions of UAV and obstacle at different times in head on conflict
numerical simulation.}
\label{Position1}
\end{figure}

\begin{figure}[ptb]
\begin{centering}
\includegraphics[scale=0.52]{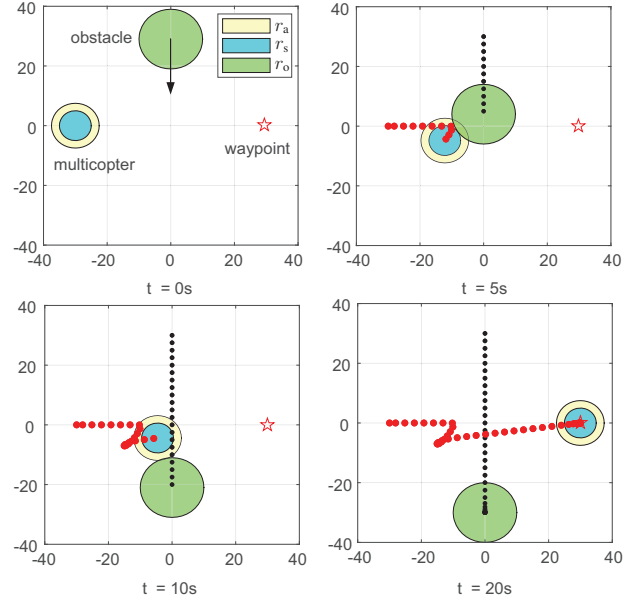}
\par\end{centering}
\caption{Positions of UAV and obstacle at different times in left converge
conflict numerical simulation.}
\label{Position1-1}
\end{figure}

\begin{figure}[ptb]
\begin{centering}
\includegraphics[scale=0.65]{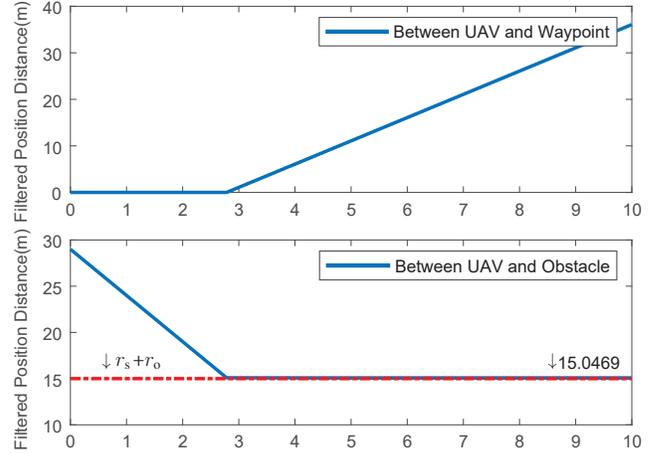}
\par\end{centering}
\caption{The filtered position distance between UAV and waypoint, obstacle
in head on conflict numerical simulation.}
\label{mindis1}
\end{figure}

As shown in Figure \ref{Position1} and Figure \ref{Position1-1},
two scenarios that one multicopter makes avoidance with one moving
non-cooperative obstacle is considered. Because the condition $v_{\text{m}}>v{_{\text{o}}}$
holds, the taking over conflict can be ignored. We simulate the head
on conflict and the converging conflict in two scenarios respectively.
In the first designed scenario, as shown in Figure \ref{Position1},
the multicopter is facing a head on conflict, where the obstacle's
velocity always points to the multicopter. The simulation parameters
are set as follows. The multicopter with the safety radius $r_{\text{s}}$
$=5$m and the avoidance radius $r_{\text{s}}$ $=7.5$m is at $\mathbf{p}\left(0\right)=\left[0~0\right]^{\text{T}}$m
initially. The waypoint $\mathbf{p}_{\text{wp}}=\mathbf{p}\left(0\right)$
is set to ensure that the multicopter is static initially. The multicopter
has the maneuver constant $l=5$, and the maximum speed $v_{\text{m}}=6\text{m/s}$.
The obstacle is at $\mathbf{p}_{\text{o}}\left(0\right)=\left[30~0\right]^{\text{T}}$m
initially with radius $r_{\text{o}}=10$m and a constant velocity
$\mathbf{v}_{\text{o}}=\left[-5~0\right]^{\text{T}}$m/s. Under the
initial conditions above and the proposed obstacle avoidance controller,
the \emph{filtered position distance }$\left\Vert \boldsymbol{\tilde{\xi}}_{\text{wp}}\left(t\right)\right\Vert $
and $\left\Vert \boldsymbol{\tilde{\xi}}{}_{\text{o}}\left(t\right)\right\Vert $
between the multicopter and the obstacle are shown in Figure \ref{mindis1}.
In this precisely constructed simulation scenario, the multicopter
cannot arrives at the waypoint, which is with probability 0 in practice.

\begin{figure}[ptb]
\begin{centering}
\includegraphics[scale=0.65]{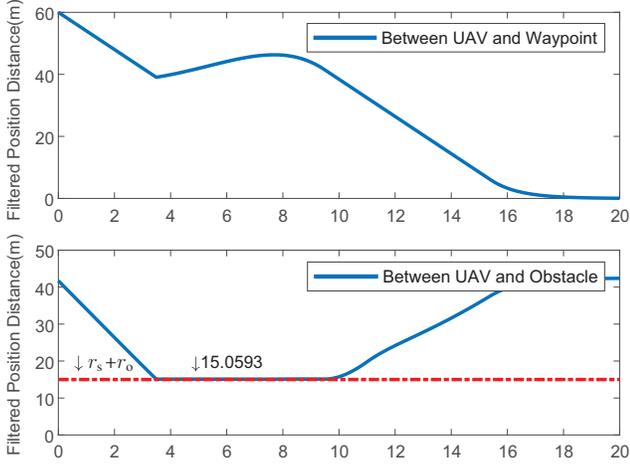}
\par\end{centering}
\caption{The filtered position distance between UAV and waypoint, obstacle
in left converge conflict numerical simulation.}
\label{mindis1-1}
\end{figure}

In the second designed scenario, as shown in Figure \ref{Position1-1},
the multicopter is facing a left converging conflict. We only change
the initial positions and obstacle's constant velocity. The multicopter
is at $\mathbf{p}\left(0\right)=\left[-30~0\right]^{\text{T}}$m initially
and the waypoint $\mathbf{p}_{\text{wp}}=\left[30~0\right]^{\text{T}}$m
is set, while the obstacle is at $\mathbf{p}_{\text{o}}\left(0\right)=\left[0~30\right]^{\text{T}}$m
initially with a constant velocity $\mathbf{v}_{\text{o}}=\left[0~-5\right]^{\text{T}}$m/s.
Under such conditions above and the proposed obstacle avoidance controller,
the \emph{filtered position distance }$\left\Vert \boldsymbol{\tilde{\xi}}_{\text{wp}}\left(t\right)\right\Vert $
and $\left\Vert \boldsymbol{\tilde{\xi}}{}_{\text{o}}\left(t\right)\right\Vert $
are shown in Figure \ref{mindis1-1}. Therefore, the multicopter can
avoid colliding the non-cooperative obstacle when facing different
types of conflicts under the proposed controller. 


\textbf{(2) Simulation with Multiple Parallel Moving Obstacles with
the Same Velocity}

\begin{figure}[ptb]
\begin{centering}
\includegraphics[scale=0.6]{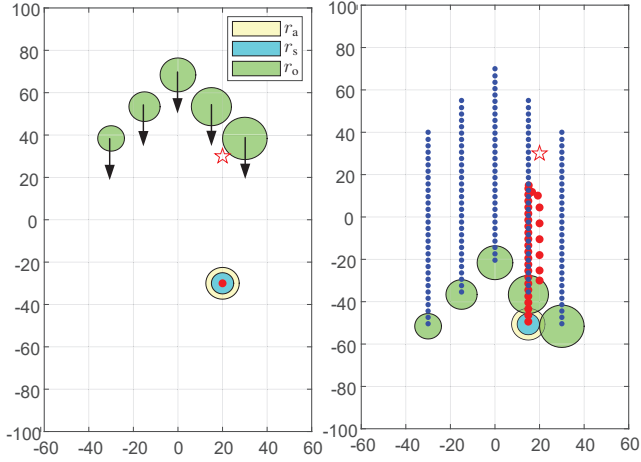}
\par\end{centering}
\caption{Positions of UAV and obstacles at different times in numerical simulation.}
\label{Position2-deadlock}
\end{figure}

\begin{figure}[ptb]
\begin{centering}
\includegraphics[scale=0.5]{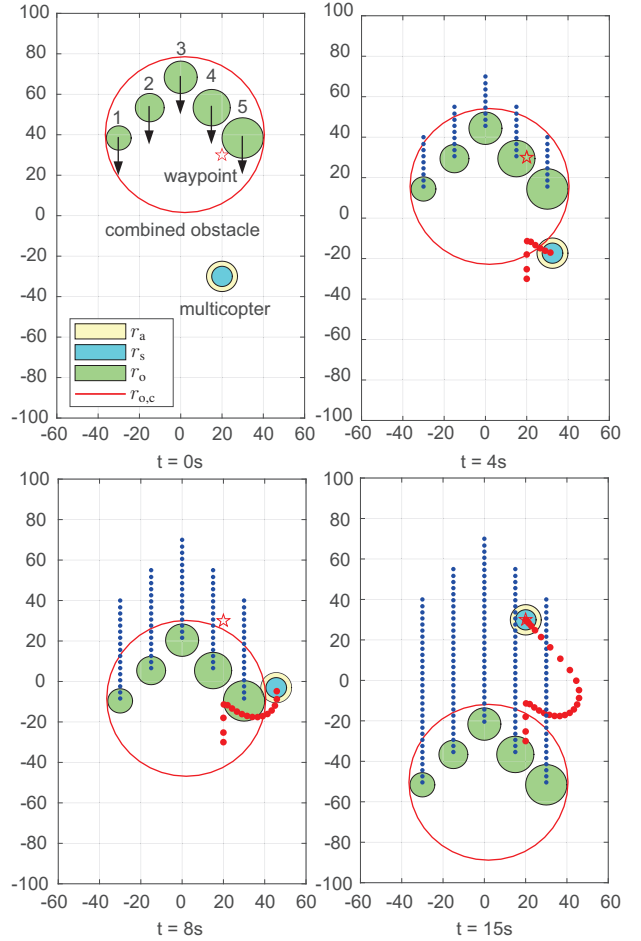} 
\par\end{centering}
\caption{Positions of UAV and combined obstacle at different times in numerical
simulation.}
\label{Position3}
\end{figure}

\begin{figure}[ptb]
\begin{centering}
\includegraphics[scale=0.55]{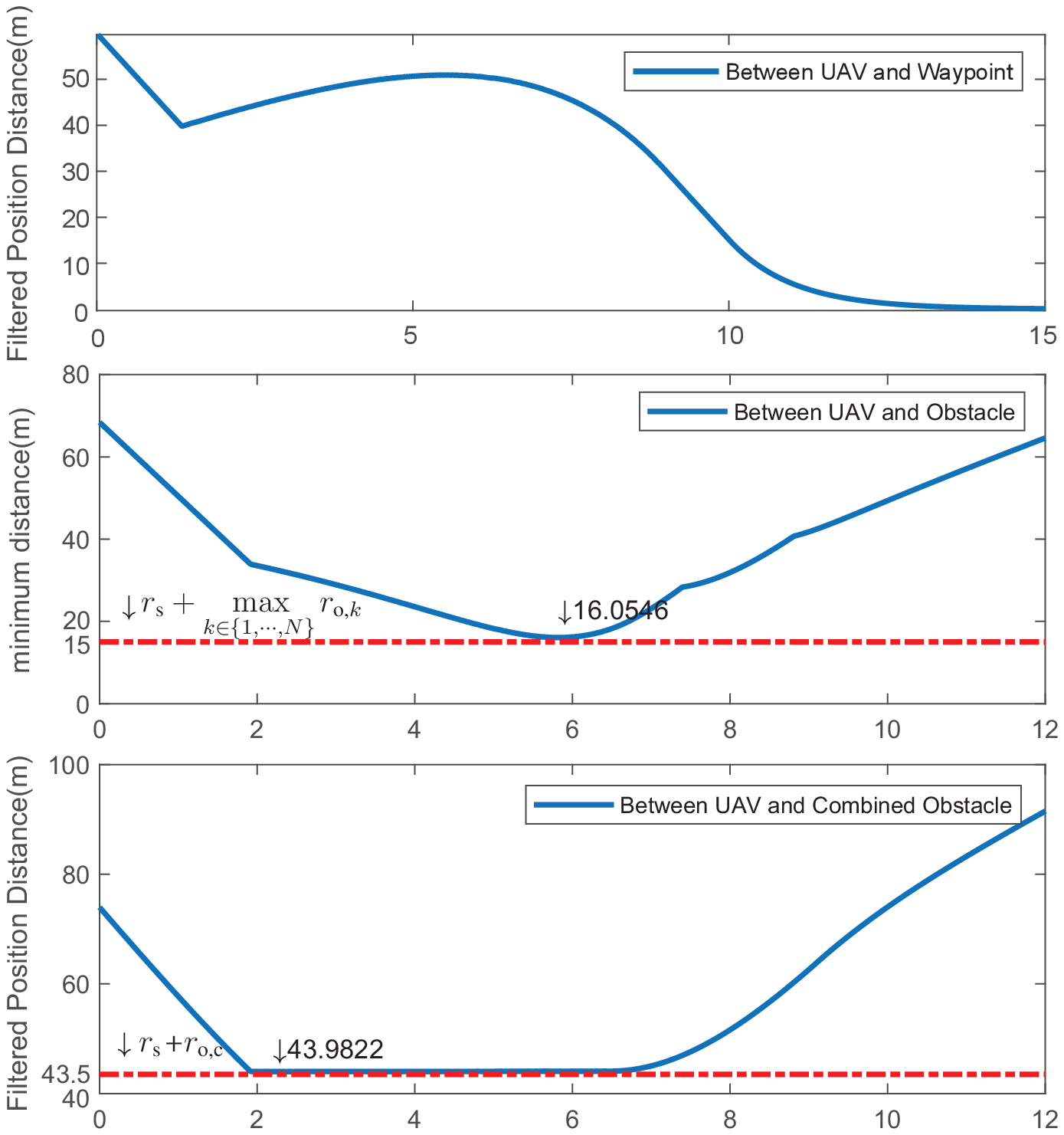} 
\par\end{centering}
\caption{The mimimum filtered position distance between UAV and waypoint, obstacles
in numerical simulation respectively.}
\label{mindis3}
\end{figure}

As shown in Figures \ref{Position2-deadlock}-\ref{Position3}, a
scenario that one static multicopter makes avoidance with $N=5$ parallel
moving non-cooperative obstacles is considered. In this scenario,
the obstacles are under homogeneous condition, similar to the horizon
wall scenario proposed in \cite{Hoekstra2001}. The simulation parameters
are set as follows. The multicopter with the safety radius $r_{\text{s}}=5$m
and the avoidance radius $r_{\text{s}}$ $=7.5$m is at $\mathbf{p}\left(0\right)=\left[20~-30\right]^{\text{T}}$m
initially. The multicopter has the maneuver constant $l=5$, the maximum
speed $v_{\text{m}}=10\text{m/s}$ and the waypoint $\mathbf{p}_{\text{wp}}=\left[20~30\right]^{\text{T}}$m.
The obstacles are at $\mathbf{p}_{\text{o},i}\left(0\right)=\left[15i-45~70-15\left\vert i-3\right\vert \right]^{\text{T}}$m
initially with radius $r_{\text{o,}i}=10+2i$ m and a constant velocity
$\mathbf{v}_{\text{o},i}=\left[0~-8\right]^{\text{T}}$m/s, $i=1,\cdots,N.$
It is worth pointing out that these parallel obstacles arrange in
the ``V''\ form with the same velocity. If we do not treat the
parallel obstacles as a combined obstacle but use a traditional controller
based on the artificial potential field method, as shown in Figure
\ref{Position2-deadlock}, the multicopter will be taken away by the
moving obstacles, unable to arrive the goal waypoint. In this scenario,
as shown in Figure \ref{Position3}, the obstacles can be taken as
one combined obstacle with $r_{\text{o,c}}=38.5$m, and \textit{Assumption
2'} is satisfied. Under the initial conditions above and the proposed
obstacle avoidance controller, the \emph{filtered position distance
}$\left\Vert \boldsymbol{\tilde{\xi}}_{\text{wp}}\left(t\right)\right\Vert $,
$\underset{i\in\left\{ 1,\cdots,5\right\} }{\min}\left\Vert \boldsymbol{\tilde{\xi}}_{\text{o},i}\left(t\right)\right\Vert $
and $\left\Vert \boldsymbol{\tilde{\xi}}_{\text{o,c}}\left(t\right)\right\Vert $
are shown in Figure \ref{mindis3}. Therefore, under the proposed
controller, the multicopter can arrive at the goal waypoint while
avoiding the parallel obstacles with the same velocity.

\textbf{(3) Simulation with Multiple Non-Parallel Moving Obstacles}

\begin{figure}[ptb]
\begin{centering}
\includegraphics[scale=0.52]{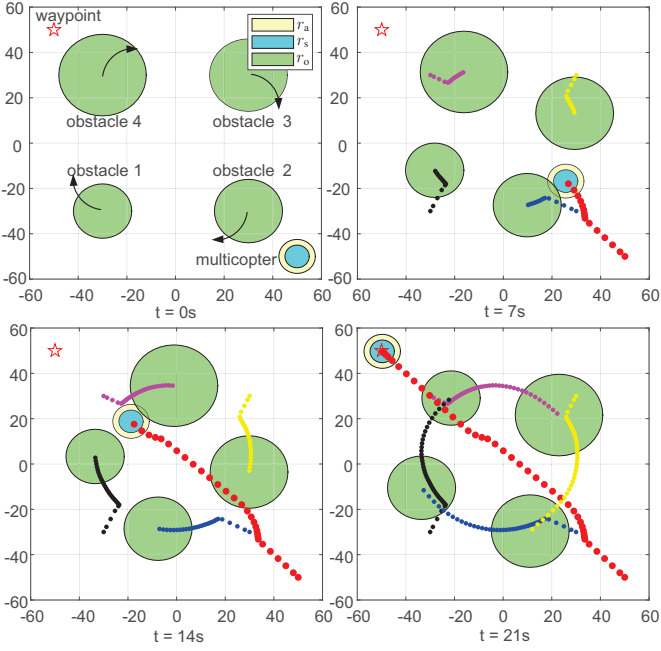}
\par\end{centering}
\caption{Positions of UAV and obstacles at different times in numerical simulation.}
\label{Position2}
\end{figure}

\begin{figure}[ptb]
\begin{centering}
\includegraphics[scale=0.52]{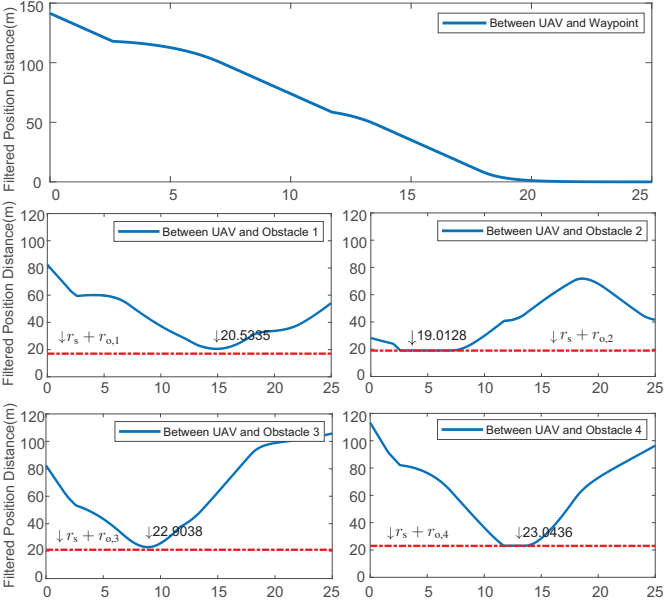}
\par\end{centering}
\caption{The filtered position distance between UAV and waypoint, each obstacle
in numerical simulation respectively.}
\label{mindis2}
\end{figure}

As shown in Figure \ref{Position2}, a scenario that one static multicopter
makes avoidance with $N=4$ non-parallel moving non-cooperative obstacles
is considered. In this scenario, the obstacles are under heterogeneous
condition. The simulation parameters are set as follows. The multicopter
with the safety radius $r_{\text{s}}=5$m and the avoidance radius
$r_{\text{s}}$ $=7.5$m is at $\mathbf{p}\left(0\right)=\left[50~-50\right]^{\text{T}}$m
initially. The multicopter has the maneuver constant $l=5$, the maximum
speed $v_{\text{m}}=9\text{m/s}$ and the waypoint $\mathbf{p}_{\text{wp}}=\left[-50~50\right]^{\text{T}}$m.
The obstacles are with radius $r_{\text{o,}i}=10+2i$ m and $v_{\text{o}}=10$m/s,
$i=1,\cdots,N.$ These obstacles have unknown trajectories but satisfy
\textit{Assumption 2''}. Under the initial conditions above and the
proposed obstacle avoidance controller, the \emph{filtered position
distance} $\left\Vert \boldsymbol{\tilde{\xi}}_{\text{wp}}\left(t\right)\right\Vert $,
$\left\Vert \boldsymbol{\tilde{\xi}}_{\text{o},i}\left(t\right)\right\Vert $
and $\left\Vert \boldsymbol{\tilde{\xi}}_{\text{o,c}}\left(0\right)\right\Vert $
are shown in Figure \ref{mindis2}, $i=1,\cdots,N.$ Therefore, under
the proposed controller, the multicopter can arrive at the goal waypoint
while avoiding the non-parallel obstacles. 

\textbf{(4) Simulation with Multiple Obstacles Using the Same Avoidance
Controller}

\begin{figure}[ptb]
\begin{centering}
\includegraphics[scale=0.5]{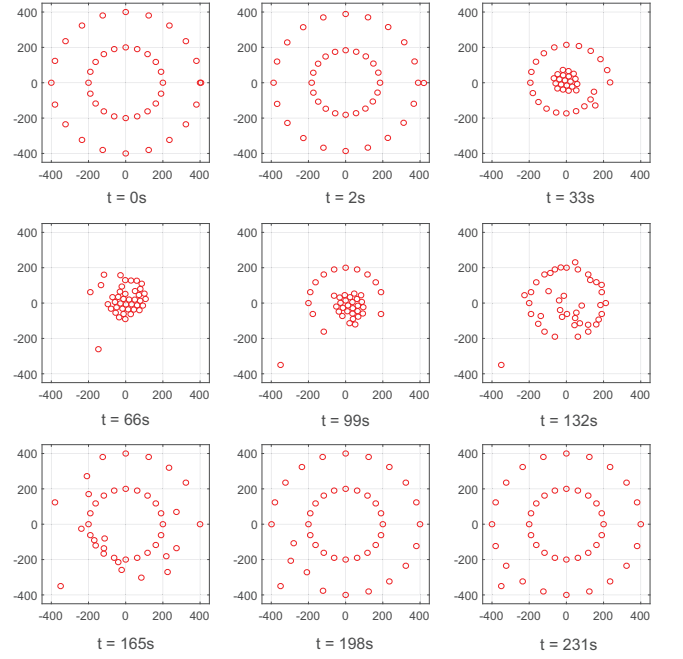}
\par\end{centering}
\caption{Positions of 41 multicopters facing a super conflicts at different
time}
\label{Position}
\end{figure}

\begin{figure}[ptb]
\begin{centering}
\includegraphics[scale=0.4]{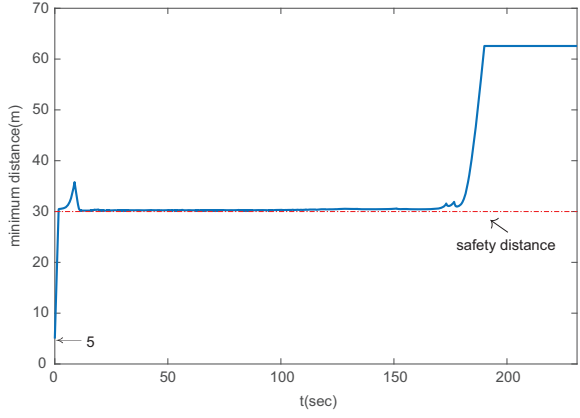}
\par\end{centering}
\caption{Mimimum distance among all multicopters facing a super conflicts}
\label{mindis}
\end{figure}

A scenario of $M=41$ multicopters with $r_{\text{s}}=15$m, $r_{\text{a}}=22.5$m
is considered, whose avoidance controllers are the same. The control
gain $l_{i}=5$ and the maximum speed of the \textit{i}th multicopter
$v_{\text{m,}i}=5+\frac{(i-1)}{8}\text{m/s},i=1,\text{···}41$. To
simulate the super conflicts proposed in \cite{Hoekstra2001}, for
the $i$th multicopter, the initial position and the waypoint are
set as 
\begin{align*}
\begin{cases}
\mathbf{p}_{i}\left(0\right) & =\left[400\cos\frac{i-1}{10}\pi~400\sin\frac{i-1}{10}\pi\right]^{\text{T}}\text{m}\\
\mathbf{p}_{\text{wp},i} & =\left[400\cos\frac{i+9}{10}\pi\text{~}400\sin\frac{i+9}{10}\pi\right]^{\text{T}}\text{m}
\end{cases},i & =1,\cdots,20\\
\begin{cases}
\mathbf{p}_{i}\left(0\right) & =\left[200\cos\frac{i-1}{10}\pi~200\sin\frac{i-1}{10}\pi\right]^{\text{T}}\text{m}\\
\mathbf{p}_{\text{wp},i} & =\left[200\cos\frac{i+9}{10}\pi\text{~}200\sin\frac{i+9}{10}\pi\right]^{\text{T}}\text{m}
\end{cases},i & =21,\cdots,40.
\end{align*}
These positions are distributed on the circumference of two circles
with center $\mathbf{o}_{1}=\mathbf{o}_{2}=\left[0~0\right]^{\text{T}}$
and radius $r_{1}=400$m, $r_{2}=200$m, respectively. For the $41$th
multicopter, its initial position is set as $\mathbf{p}_{41}\left(0\right)=\left[405~0\right]^{\text{T}}$m,
which is closed to the 1st multicopter initially; the waypoint is
$\mathbf{p}_{\text{wp},41}=\left[-350~-350\right]^{\text{T}}\text{m}$.
This is to simulate the situation a multicopter appears in another's
safety area accidentally. As shown in Figures \ref{Position}-\ref{mindis},
each multicopter can fly to its waypoint without deadlock and conflict
with other multicopters. The minimum distance among multicopters is
shown in Figure \ref{mindis}. The minimum distance among multicopters
is increased rapidly. At about $t=2$s, the conflict between the 1st
and the 41th multicopters has disappeared as soon as possible. From
then on, no conflict happens again. The result shows that the multicopter
can arrive at the goal waypoint while avoiding the obstacles under
the proposed controller when the multicopter and the obstacles have
the same avoidance controller. 

\subsection{Flight Experiment}

A motion capture system called OptiTrack is installed, from which
we can get the ground truth of the position, velocity and orientation
of each multicopter. The laptop is connected to these multicopters
and OptiTrack by a local network, providing the proposed controller
and a real-time position plotting module. In the two experiments,
our design is similar to the last two simulation scenarios, assuming
that $r_{\text{s}}=0.2\text{{m}},r_{\text{a}}=0.25\text{{m}},r_{\text{o}}=0.2\text{{m}}.$
The effectiveness of the proposed controller and the correctiveness
of the result extension to multiple moving obstacles are further verified
by the experiments.

\textbf{(1) Multiple Parallel Moving Obstacles with Different Directions}

\begin{figure}[ptb]
\begin{centering}
\includegraphics[scale=0.2]{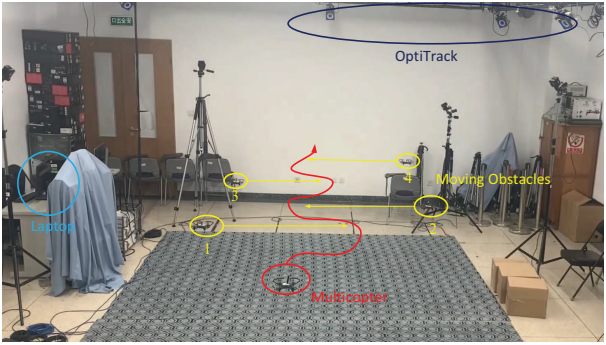} 
\par\end{centering}
\caption{Multiple moving obstacles with different directions.}
\label{sceneraio}
\end{figure}

\begin{figure}[ptb]
\begin{centering}
\includegraphics[scale=0.45]{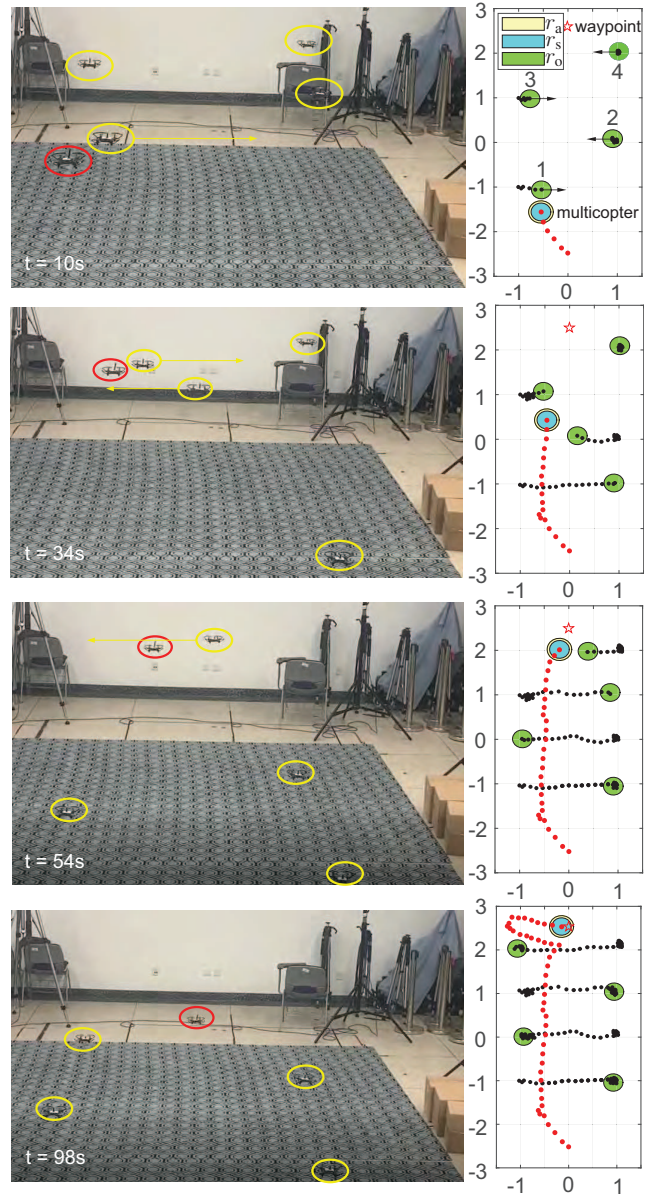} 
\par\end{centering}
\caption{Positions of UAV and obstacles in the case that multiple parallel
moving obstacles with different directions.}
\label{flightpos}
\end{figure}

As shown in Figure \ref{sceneraio}, the sceneraio contains one multicopter
and $N$ = 4 moving obstacles. The initial position and the waypoint
of multicopter is set as $\mathbf{p}\left(0\right)=\left[0~-2.5\right]^{\text{T}}$
and $\mathbf{p}_{\text{wp}}=\left[0~2.5\right]^{\text{T}}$with $v_{\text{m}}=0.12$m/s.
The initial position of the $i$th obstacle $\mathbf{p}_{\text{o},i}\left(0\right)=\left[\left(-1\right)^{i}~i-2\right]^{\text{T}}$m
and the velocity $\mathbf{v}{}_{\text{o},i}=\left[0.1\left(-1\right)^{i+1}~0\right]^{\text{T}}$m/s
is set, $i=1,\cdots,N.$ These obstacles satisfy \textit{Assumptions
2'}. It is worth noting that the velocity direction for each obstacle
is opposite to its neighboring obstacle. Therefore, similar to multiple
non-parallel moving obstacles, the situation can be degraded to the
one moving obstacle avoidance control problem. Finally, the multicopter
can complete its route at 98s, keeping a safe distance from moving
obstacles without conflict. The position of multicopter and obstacles
during the whole flight experiment is shown in Figure \ref{flightpos}.

\textbf{(2) Multiple Parallel Moving Obstacles with the Same Velocity}

\begin{figure}[ptb]
\begin{centering}
\includegraphics[scale=0.5]{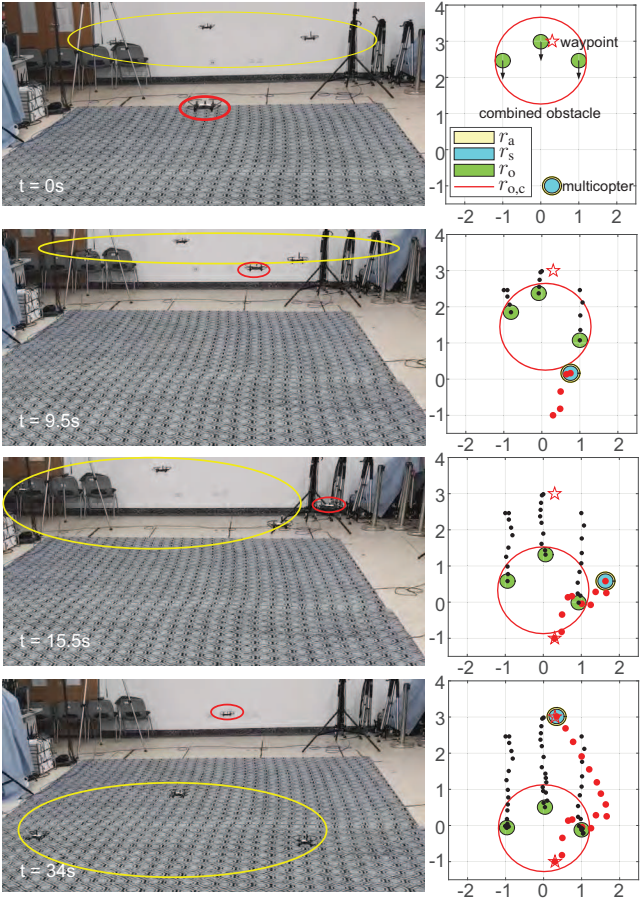} 
\par\end{centering}
\caption{Positions of UAV and obstacles in the case that multiple parallel
moving obstacles with the same velocity.}
\label{flightpos2}
\end{figure}

As shown in Figure \ref{flightpos2}, the sceneraio contains one multicopter
and $N$ = 3 moving obstacles. The initial position and the waypoint
of multicopter is set as $\mathbf{p}\left(0\right)=\left[0.3~-1\right]^{\text{T}}$
and $\mathbf{p}_{\text{\text{wp}}}=\left[0.3~3\right]^{\text{T}}$with
$v_{\text{m}}=0.12$m/s. The initial position and the velocity of
the obstacle $\mathbf{p}_{\text{o},i}=\left[2-i~3-0.5\left\vert i-2\right\vert \right]^{\text{T}}$m,
$\mathbf{v}{}_{\text{o},i}=\left[0~-0.1\right]^{\text{T}}$m/s is
set, $i=1,\cdots,N.$ Similar to the simulation, these parallel obstacles
arrange in a ``V'' shape. Under these conditions, these obstacles
can be regard as a combined obstacle while \textit{Assumptions 2''}
is satisfied. Finally, the multicopter can complete its route at 34s,
keeping a safe distance from these parallel moving obstacles without
confliction. The position of multicopter and obstacles is shown in
Figure \ref{flightpos2}.

\section{Conclusions}

The moving non-cooperative obstacle avoidance problem is studied in
this paper. First, a multicopter model and obstacle model are introduced.
Since the multicopter's speed is confined, the necessary condition
is to limit the velocity of non-cooperative moving obstacles. The
force field method is used to solve the problem. During the controller
design process, Lyapunov-like functions are designed with formal analysis
and proofs, the instability about angle rather than position is proved
for the case that the multicopter is in front of an obstacle moving
direction, showing that one moving obstacle avoidance control problem
can be solved. Furthermore, one moving obstacle avoidance control
extends to two types of multiple moving obstacle avoidance control
problems. Simulations and experiments are given to show the effectiveness
of the proposed method from the functional requirement and the safety
requirement. In the future, more studies are deserved to spend on
developing strategies to deal with complex multiple moving obstacles.

\section{Appendix}

\subsection{Proof of Proposition 1}

First, we have 
\[
\mathbf{v}^{\text{T}}\mathbf{\dot{v}}=-l\mathbf{v}^{\text{T}}\mathbf{v}+l\mathbf{v}^{\text{T}}\text{sa}{\text{t}}\left(\mathbf{v}_{\text{c}},{v_{\text{m}}}\right){.}
\]
This implies 
\[
\frac{1}{2}\frac{\text{d}\left\Vert \mathbf{v}\right\Vert ^{\text{2}}}{\text{d}t}=-l\left\Vert \mathbf{v}\right\Vert ^{\text{2}}+l\mathbf{v}^{\text{T}}\text{sa}{\text{t}}\left(\mathbf{v}_{\text{c}},{v_{\text{m}}}\right){.}
\]
Consequently, 
\begin{equation}
\frac{\text{d}\left\Vert \mathbf{v}\right\Vert }{\text{d}t}=-l\left\Vert \mathbf{v}\right\Vert +l\frac{1}{\left\Vert \mathbf{v}\right\Vert }\mathbf{v}^{\text{T}}\text{sa}{\text{t}}\left(\mathbf{v}_{\text{c}},{v_{\text{m}}}\right){.}\label{w}
\end{equation}
The solution to (\ref{w}) is 
\[
\left\Vert \mathbf{v}\left(t\right)\right\Vert =e^{-lt}\left\Vert \mathbf{v}\left(0\right)\right\Vert +
{\displaystyle \int\nolimits _{0}^{t}}
e^{-l\left(t-s\right)}\frac{l}{\left\Vert \mathbf{v}\right\Vert }\mathbf{v}^{\text{T}}\text{sa}{\text{t}}\left(\mathbf{v}_{\text{c}},{v_{\text{m}}}\right)\text{d}s.
\]
Since $\left\Vert \text{sa}{\text{t}}\left(\mathbf{v}_{\text{c}},{v_{\text{m}}}\right)\right\Vert \leq{v_{\text{m}}}$,
we have 
\begin{align*}
\left\Vert \mathbf{v}\left(t\right)\right\Vert  & \leq e^{-lt}\left\Vert \mathbf{v}\left(0\right)\right\Vert +
{\displaystyle \int\nolimits _{0}^{t}}
e^{-l\left(t-s\right)}l{v_{\text{m}}}\text{d}s\\
 & =e^{-lt}\left\Vert \mathbf{v}\left(0\right)\right\Vert +{v_{\text{m}}-}e^{-lt}{v_{\text{m}}}.
\end{align*}
With $\left\Vert \mathbf{v}\left(0\right)\right\Vert \leq{v_{\text{m}},}$
we have $\left\Vert \mathbf{v}\left(t\right)\right\Vert \leq{v_{\text{m}},}$
$t\geq0{.}$

\subsection{Proof of Proposition 2}

\textbf{(i) Proof of sufficiency}. Let 
\begin{align*}
p & =\mathbf{\tilde{p}}_{\text{o}}^{\text{T}}\mathbf{\tilde{p}}{_{\text{o}}}\\
\delta & =\boldsymbol{\tilde{\xi}}{_{\text{o}}^{\text{T}}}\boldsymbol{\tilde{\xi}}{_{\text{o}}}-\frac{1}{l^{2}}\mathbf{\tilde{v}}{_{\text{o}}^{\text{T}}\mathbf{\tilde{v}}{_{\text{o}}}.}
\end{align*}
According to (\ref{errors}), we have 
\begin{align}
\boldsymbol{\tilde{\xi}}{_{\text{o}}^{\text{T}}}\boldsymbol{\tilde{\xi}}{_{\text{o}}} & =\left(\mathbf{\tilde{p}}{_{\text{o}}}+\frac{1}{l}\mathbf{\tilde{v}}{_{\text{o}}}\right)^{\text{T}}\left(\mathbf{\tilde{p}}{_{\text{o}}}+\frac{1}{l}\mathbf{\tilde{v}}{_{\text{o}}}\right)\nonumber \\
 & =\mathbf{\tilde{p}}_{\text{o}}^{\text{T}}\mathbf{\tilde{p}}{_{\text{o}}}+\frac{1}{l^{2}}\mathbf{\tilde{v}}{_{\text{o}}^{\text{T}}\mathbf{\tilde{v}}{_{\text{o}}}}+\frac{2}{l}\mathbf{\tilde{v}}{_{\text{o}}^{\text{T}}\mathbf{\tilde{p}}{_{\text{o}}}.}\label{filtererror2}
\end{align}
Since 
\[
\frac{\text{d}p}{\text{d}t}=2\mathbf{\tilde{p}}_{\text{o}}^{\text{T}}{\mathbf{\tilde{v}}{_{\text{o}}}}
\]
using the equation (\ref{filtererror2}), we further have 
\begin{equation}
\frac{\text{d}p}{\text{d}t}=-lp+l\delta.\label{equ}
\end{equation}
The solution $p\left(t\right)$ can be expressed as 
\begin{equation}
p\left(t\right)=e^{-lt}p\left(0\right)+
{\displaystyle \int\nolimits _{0}^{t}}
e^{-l\left(t-s\right)}l\delta\left(s\right)\text{d}s.\label{z}
\end{equation}
With (\ref{rv1}) in hand, if condition (\ref{p3condition}) is satisfied,
then 
\[
\delta\left(t\right)=\boldsymbol{\tilde{\xi}}_{\text{o}}^{\text{T}}\boldsymbol{\tilde{\xi}}_{\text{o}}-\frac{1}{l^{2}}\mathbf{\tilde{v}}{_{\text{o}}^{\text{T}}\mathbf{\tilde{v}}{_{\text{o}}}}\geq{r^{2}.}
\]
Since $\left\Vert \mathbf{\tilde{p}}{_{\text{o}}}\left(0\right)\right\Vert >r,$
we have $p\left(0\right)>r^{2}.$ The solution in (\ref{z}) satisfies
\begin{align*}
p\left(t\right) & \geq e^{-lt}r^{2}+{r^{2}}
{\displaystyle \int\nolimits _{0}^{t}}
e^{-l\left(t-s\right)}l\text{d}s\\
 & =r^{2}.
\end{align*}
Based on it, we have $\left\Vert \mathbf{\tilde{p}}{_{\text{o}}\left(t\right)}\right\Vert \geq r,$
where $t\geq0$. If $\frac{\mathbf{v}^{\text{T}}{{\mathbf{v}}_{\text{o}}}}{\left\Vert \mathbf{v}\right\Vert \left\Vert {{\mathbf{v}}_{\text{o}}}\right\Vert }=-1,$
then the multicopter and the obstacle are in the case shown in Figure
1(b). Thus, 
\[
\frac{1}{l^{2}}\mathbf{\tilde{v}}{_{\text{o}}^{\text{T}}\mathbf{\tilde{v}}{_{\text{o}}}}=r_{\text{v}}.
\]
Consequently, $\delta\left(t\right)={r^{2}.}$ Then, if $\left\Vert \mathbf{\tilde{p}}{_{\text{o}}\left(0\right)}\right\Vert =r,$
then $\left\Vert \mathbf{\tilde{p}}{_{\text{o}}\left(t\right)}\right\Vert \equiv r.$
Furthermore, if $\left\Vert \boldsymbol{\tilde{\xi}}{_{\text{o}}}\left(t\right)\right\Vert >\sqrt{r^{2}+r_{\text{v}}^{2}}$
and $\left\Vert \mathbf{\tilde{p}}{_{\text{o}}}\left(0\right)\right\Vert >r,$
then $\left\Vert \mathbf{\tilde{p}}{_{\text{o}}}\left(t\right)\right\Vert >r,$
where $t>0$.

\textbf{(ii) Proof of necessity}. Given any $\epsilon_{\text{o}}>0,$
we will show if 
\begin{equation}
\left\Vert \boldsymbol{\tilde{\xi}}{_{\text{o}}}\left(t\right)\right\Vert ^{2}=r^{2}+r_{\text{v}}^{2}-\epsilon_{\text{o}},
\end{equation}
and $\left\Vert \mathbf{\tilde{p}}{_{\text{o}}}\left(0\right)\right\Vert =r,$
then there exists a case that $\left\Vert \mathbf{\tilde{p}}{_{\text{o}}}\left(t\right)\right\Vert <r,$
where $t\geq0$. Consider a case $\frac{\mathbf{v}^{\text{T}}{{\mathbf{v}}_{\text{o}}}}{\left\Vert \mathbf{v}\right\Vert \left\Vert {{\mathbf{v}}_{\text{o}}}\right\Vert }=-1.$
Then the multicopter and the obstacle are in the case shown in Figure
1(b). Thus, 
\[
\delta\left(t\right)=\boldsymbol{\tilde{\xi}}_{\text{o}}^{\text{T}}\boldsymbol{\tilde{\xi}}_{\text{o}}-\frac{1}{l^{2}}\mathbf{\tilde{v}}{_{\text{o}}^{\text{T}}\mathbf{\tilde{v}}{_{\text{o}}}}={r^{2}-\epsilon_{\text{o}}.}
\]
According to (\ref{z}), we have 
\[
p\left(t\right)=r^{2}-
{\displaystyle \int\nolimits _{0}^{t}}
e^{-l\left(t-s\right)}l\epsilon_{\text{o}}\text{d}s.
\]
Therefore, $\left\Vert \mathbf{\tilde{p}}{_{\text{o}}}\left(t\right)\right\Vert <r,$
where $t\geq0$.

\subsection{Proof of Lemma 2}

First, we have 
\begin{equation}
\boldsymbol{\dot{\tilde{\xi}}}{_{\text{o}}}=\mathbf{-}\text{sa}{\text{t}}\left(\text{sa}{\text{t}}\left(k_{1}\boldsymbol{\tilde{\xi}}{_{\text{wp}}},v_{\text{m}}\right)-a{_{\text{o}}}\boldsymbol{\tilde{\xi}}{_{\text{o}}},{v_{\text{m}}}\right)-\mathbf{a}_{\text{o}}.\label{errorobstacle}
\end{equation}
Since 
\[
\frac{\text{d}\boldsymbol{\tilde{\xi}}_{\text{o}}^{\text{T}}\boldsymbol{\tilde{\xi}}_{\text{o}}}{\text{d}t}=2\boldsymbol{\tilde{\xi}}_{\text{o}}^{\text{T}}\boldsymbol{\dot{\tilde{\xi}}}{_{\text{o}}}
\]
by using (\ref{errorobstacle}), we have 
\begin{equation}
\frac{\text{d}\boldsymbol{\tilde{\xi}}_{\text{o}}^{\text{T}}\boldsymbol{\tilde{\xi}}_{\text{o}}}{\text{d}t}=\mathbf{-}2\boldsymbol{\tilde{\xi}}_{\text{o}}^{\text{T}}\text{sa}{\text{t}}\left(\text{sa}{\text{t}}\left(k_{1}\boldsymbol{\tilde{\xi}}{_{\text{wp}}},v_{\text{m}}\right)-a{_{\text{o}}}\boldsymbol{\tilde{\xi}}{_{\text{o}}},{v_{\text{m}}}\right)-2\boldsymbol{\tilde{\xi}}_{\text{o}}^{\text{T}}\mathbf{a}_{\text{o}}.\label{error2}
\end{equation}
We consider a case that, at time $t=t_{1}>0,$ suppose 
\begin{equation}
\left\Vert \boldsymbol{\tilde{\xi}}_{\text{o}}\left(t_{1}\right)\right\Vert =\left(1-\epsilon_{\gamma}\right)\gamma r_{\text{s}}+{{r}_{\text{o}}}\label{acase}
\end{equation}
where $0<\epsilon_{\gamma}<1-\frac{1}{\gamma}$ satisfies 
\[
1<\left(1-\epsilon_{\gamma}\right)\gamma<\gamma.
\]
This implies no conflict at time $t_{1}.$ In this case, according
to (\ref{Property}), we have 
\[
V_{\text{o}}\left(t_{1}\right)=\frac{k_{2}}{\epsilon\left\Vert \boldsymbol{\tilde{\xi}}{_{\text{o}}}\left(t_{1}\right)\right\Vert }+g_{1}\left(\epsilon_{\text{s}}\right)
\]
where $g_{1}\left(\epsilon_{\text{s}}\right)$ is a term related to
$\epsilon_{\text{s}}>0$ which satisfies $g_{1}\left(\epsilon_{\text{s}}\right)\rightarrow0$
as $\epsilon_{\text{s}}\rightarrow0.$ As a result, according to definition
of $a_{\text{o}},$ 
\[
a{_{\text{o}}}\boldsymbol{\tilde{\xi}}{_{\text{o}}}\left(t_{1}\right)=\left(\frac{k_{2}}{\epsilon}\frac{1}{\left\Vert \boldsymbol{\tilde{\xi}}{_{\text{o}}}\left(t_{1}\right)\right\Vert ^{3}}+g_{2}\left(\epsilon_{\text{s}}\right)\right)\boldsymbol{\tilde{\xi}}{}_{\text{o}}\left(t_{1}\right)
\]
where $g_{2}\left(\epsilon_{\text{s}}\right)$ is a term related to
$\epsilon_{\text{s}}>0$ satisfy $g_{2}\left(\epsilon_{\text{s}}\right)\rightarrow0$
as $\epsilon_{\text{s}}\rightarrow0.$ Since $\epsilon$ is chosen
very small, $a_{\text{o}}\boldsymbol{\tilde{\xi}}{}_{\text{o}}$ dominates
the term sa${\text{t}}\left(k_{1}\boldsymbol{\tilde{\xi}}{_{\text{wp}}},v_{\text{m}}\right)-a{_{\text{o}}}\boldsymbol{\tilde{\xi}}{_{\text{o}}}$.
Therefore, at time $t=t_{1},$ (\ref{error2}) becomes 
\begin{align}
\frac{\text{d}\boldsymbol{\tilde{\xi}}_{\text{o}}^{\text{T}}\boldsymbol{\tilde{\xi}}_{\text{o}}}{\text{d}t}\geq & 2\boldsymbol{\tilde{\xi}}_{\text{o}}^{\text{T}}\text{sa}{\text{t}}\left(\frac{k_{2}}{\epsilon}\frac{1}{\left\Vert \boldsymbol{\tilde{\xi}}{_{\text{o}}}\right\Vert ^{3}}\boldsymbol{\tilde{\xi}}{_{\text{o}}},{v_{\text{m}}}\right)-2\boldsymbol{\tilde{\xi}}_{\text{o}}^{\text{T}}\mathbf{a}_{\text{o}}\nonumber \\
 & -\left\vert g_{3}\left(\epsilon_{\text{s}}\right)\right\vert -\left\vert g_{4}\left(\epsilon\right)\right\vert \nonumber \\
\geq & 2\left\Vert \boldsymbol{\tilde{\xi}}_{\text{o}}\right\Vert {v_{\text{m}}-}2\left\Vert \boldsymbol{\tilde{\xi}}_{\text{o}}\right\Vert \left\Vert \mathbf{a}_{\text{o}}\right\Vert -\left\vert g_{3}\left(\epsilon_{\text{s}}\right)\right\vert -\left\vert g_{4}\left(\epsilon\right)\right\vert \label{diffeeror}
\end{align}
where $g_{3}\left(\epsilon_{\text{s}}\right)$ is a term related to
$\epsilon_{\text{s}}>0$ which satisfies $g_{3}\left(\epsilon_{\text{s}}\right)\rightarrow0$
as $\epsilon_{\text{s}}\rightarrow0,$ and $g_{4}\left(\epsilon\right)$
is a term related to $\epsilon>0$ which satisfies $g_{4}\left(\epsilon\right)\rightarrow0$
as $\epsilon\rightarrow0.$ Since 
\[
\frac{\text{d}\boldsymbol{\tilde{\xi}}_{\text{o}}^{\text{T}}\boldsymbol{\tilde{\xi}}_{\text{o}}}{\text{d}t}=\frac{\text{d}\left\Vert \boldsymbol{\tilde{\xi}}_{\text{o}}\right\Vert ^{2}}{\text{d}t}=2\left\Vert \boldsymbol{\tilde{\xi}}_{\text{o}}\right\Vert \frac{\text{d}\left\Vert \boldsymbol{\tilde{\xi}}_{\text{o}}\right\Vert }{\text{d}t}
\]
and ${v_{\text{m}}}>v_{\text{o}},$ by using (\ref{diffeeror}), there
exist sufficiently small $\epsilon_{\text{s}},\epsilon>0$ such that
\begin{align*}
\left.\frac{\text{d}\left\Vert \boldsymbol{\tilde{\xi}}_{\text{o}}\right\Vert }{\text{d}t}\right\vert _{t=t_{1}}\geq & 2{v_{\text{m}}-}2v_{\text{o}}-\frac{1}{\left(1-\epsilon_{\gamma}\right)\gamma r_{\text{s}}+{{r}_{\text{o}}}}\left\vert g_{3}\left(\epsilon_{\text{s}}\right)\right\vert \\
 & -\frac{1}{\left(1-\epsilon_{\gamma}\right)\gamma r_{\text{s}}+{{r}_{\text{o}}}}\left\vert g_{4}\left(\epsilon\right)\right\vert .
\end{align*}
There exist sufficiently small $\epsilon_{\text{s}},\epsilon$ such
that 
\begin{align*}
 & 2{v_{\text{m}}-}2v_{\text{o}}-\frac{1}{\left(1-\epsilon_{\gamma}\right)\gamma r_{\text{s}}+{{r}_{\text{o}}}}\left\vert g_{3}\left(\epsilon_{\text{s}}\right)\right\vert \\
-\frac{1}{\left(1-\epsilon_{\gamma}\right)\gamma r_{\text{s}}+{{r}_{\text{o}}}}\left\vert g_{4}\left(\epsilon\right)\right\vert  & >0.
\end{align*}
This implies that if the filtered position error to the obstacle for
the multicopter satisfies (\ref{acase}), the filtered position error
will not be decreased any more. From the case, the filtered position
error cannot be smaller than $\left(1-\epsilon_{\gamma}\right)\gamma r_{\text{s}}+{{r}_{\text{o}}}$
anymore. Therefore, by using the fact $\left(1-\epsilon_{\gamma}\right)\gamma r_{\text{s}}+{{r}_{\text{o}}}>r_{\text{s}}+{{r}_{\text{o}}}$,
we can claim $\left\Vert \boldsymbol{\tilde{\xi}}_{\text{o}}\right\Vert >r_{\text{s}}+{{r}_{\text{o}}.}$

\subsection{Proof of Lemma 3}
\begin{itemize}
\item \textbf{Step 1. A Lyapunov function defined. }Define 
\[
V_{1,1}=1+\frac{\boldsymbol{\tilde{\xi}}_{\text{o}}^{\text{T}}\mathbf{v}{_{\text{o}}}}{\left\Vert \mathbf{v}{_{\text{o}}}\right\Vert \left\Vert \boldsymbol{\tilde{\xi}}{_{\text{o}}}\right\Vert }.
\]
According to (\ref{costheta}), we have 
\begin{align*}
V_{1,1} & =1+\cos\theta\\
 & =2\cos^{2}\frac{\theta}{2}.
\end{align*}
\item \textbf{Step 2. Derivative of Lyapunov function. }The derivative of
$V_{1,1}$ along the solution to (\ref{obmodel}) is 
\begin{align}
\dot{V}_{1,1} & =\frac{1}{\left\Vert \mathbf{v}{_{\text{o}}}\right\Vert \left\Vert \boldsymbol{\tilde{\xi}}{_{\text{o}}}\right\Vert }\boldsymbol{\dot{\tilde{\xi}}}_{\text{o}}^{\text{T}}{\mathbf{P}_{\text{o}}}\mathbf{v}{_{\text{o}}}\nonumber \\
 & =\frac{1}{\left\Vert \mathbf{v}{_{\text{o}}}\right\Vert \left\Vert \boldsymbol{\tilde{\xi}}{_{\text{o}}}\right\Vert }\left(\mathbf{-}\text{sa}{\text{t}}\left(\text{sa}{\text{t}}\left(k_{1}\boldsymbol{\tilde{\xi}}{_{\text{wp}}},v_{\text{m}}\right)\right.\right.\nonumber \\
 & \left.\left.-a{_{\text{o}}}\boldsymbol{\tilde{\xi}}{_{\text{o}}},{v_{\text{m}}}\right)-\mathbf{v}{}_{\text{o}}\right)^{\text{{T}}}{\mathbf{P}_{\text{o}}}\mathbf{v}{_{\text{o}}}\label{dV111}
\end{align}
where 
\[
{\mathbf{P}_{\text{o}}}=\mathbf{I}_{2}-\frac{\boldsymbol{\tilde{\xi}}{_{\text{o}}}\boldsymbol{\tilde{\xi}}_{\text{o}}^{\text{T}}}{\left\Vert \boldsymbol{\tilde{\xi}}{_{\text{o}}}\right\Vert ^{2}}.
\]
By the definition of sa${\text{t}}\left(\cdot\right),$ we can rewrite
the controller (\ref{control_p2_1}) as 
\[
{\mathbf{v}{_{\text{c}}}}=-{{\kappa}_{{v_{\text{m,1}}}}}\boldsymbol{\tilde{\xi}}{_{\text{wp}}}+{{\kappa}_{{v_{\text{m,2}}}}}a{_{\text{o}}}\boldsymbol{\tilde{\xi}}{_{\text{o}}}
\]
where ${{\kappa}_{{v_{\text{m,1}}}},{\kappa}_{{v_{\text{m,2}}}}>0.}$
Then $\dot{V}_{1,1}$ in (\ref{dV111}) becomes 
\begin{align*}
\dot{V}_{1,1} & =-\frac{1}{\left\Vert \mathbf{v}{_{\text{o}}}\right\Vert \left\Vert \boldsymbol{\tilde{\xi}}{_{\text{o}}}\right\Vert }\mathbf{v}_{\text{o}}^{\text{T}}{\mathbf{P}_{\text{o}}}\mathbf{v}{_{\text{o}}}-\frac{{{\kappa}_{{v_{\text{m,1}}}}}}{\left\Vert \mathbf{v}{_{\text{o}}}\right\Vert \left\Vert \boldsymbol{\tilde{\xi}}{_{\text{o}}}\right\Vert }\boldsymbol{\tilde{\xi}}_{\text{wp}}^{\text{T}}{\mathbf{P}_{\text{o}}}\mathbf{v}{_{\text{o}}}\\
 & \text{ \ \ }+\frac{{{\kappa}_{{v_{\text{m,2}}}}}a{_{\text{o}}}}{\left\Vert \mathbf{v}{_{\text{o}}}\right\Vert \left\Vert \boldsymbol{\tilde{\xi}}{_{\text{o}}}\right\Vert }\boldsymbol{\tilde{\xi}}_{\text{o}}^{\text{T}}{\mathbf{P}_{\text{o}}}\mathbf{v}{_{\text{o}}.}
\end{align*}
Since $\boldsymbol{\tilde{\xi}}_{\text{o}}^{\text{T}}{\mathbf{P}_{\text{o}}}\equiv\mathbf{0\ }$according
to the definition of ${\mathbf{P}_{\text{o}}},$ the derivative $\dot{V}_{1,1}$
further becomes 
\[
\dot{V}_{1,1}=-\frac{1}{\left\Vert \mathbf{v}{_{\text{o}}}\right\Vert \left\Vert \boldsymbol{\tilde{\xi}}{_{\text{o}}}\right\Vert }\mathbf{v}_{\text{o}}^{\text{T}}{\mathbf{P}_{\text{o}}}\mathbf{v}{_{\text{o}}}-\frac{{{\kappa}_{{v_{\text{m,1}}}}}}{\left\Vert \mathbf{v}{_{\text{o}}}\right\Vert \left\Vert \boldsymbol{\tilde{\xi}}{_{\text{o}}}\right\Vert }\boldsymbol{\tilde{\xi}}_{\text{wp}}^{\text{T}}{\mathbf{P}_{\text{o}}}\mathbf{v}{_{\text{o}}.}
\]
Since $\mathbf{v}{_{\text{o}}\neq0}${ and }the goal waypoint is
static, then 
\[
\mathbf{p}{_{\text{o}}}-\mathbf{p}{_{\text{wp}}}=\mathbf{p}{_{\text{o}}}\left(0\right)+\mathbf{v}{_{\text{o}}t}-\mathbf{p}{_{\text{wp}}.}
\]
Consequently, 
\begin{align*}
\boldsymbol{\tilde{\xi}}_{\text{wp}}^{\text{T}}{\mathbf{P}_{\text{o}}}\mathbf{v}{_{\text{o}}} & =\left(\boldsymbol{\tilde{\xi}}{_{\text{o}}}+\left(\mathbf{p}{_{\text{o}}}-\mathbf{p}{_{\text{wp}}}\right)\right)^{\text{T}}{\mathbf{P}_{\text{o}}}\mathbf{v}{_{\text{o}}}\\
 & =\boldsymbol{\tilde{\xi}}_{\text{o}}^{\text{T}}{\mathbf{P}_{\text{o}}}\mathbf{v}{_{\text{o}}}+\left(\mathbf{p}{_{\text{o}}}-\mathbf{p}{_{\text{wp}}}\right)^{\text{T}}{\mathbf{P}_{\text{o}}}\mathbf{v}{_{\text{o}}}\\
 & =\left(\mathbf{p}{_{\text{o}}}-\mathbf{p}{_{\text{wp}}}\right)^{\text{T}}{\mathbf{P}_{\text{o}}}\mathbf{v}{_{\text{o}}}\\
 & =\left(\mathbf{p}{_{\text{o}}}\left(0\right)-\mathbf{p}{_{\text{wp}}}\right)^{\text{T}}{\mathbf{P}_{\text{o}}}\mathbf{v}{_{\text{o}}}+{t}\mathbf{v}_{\text{o}}^{\text{T}}{\mathbf{P}_{\text{o}}}\mathbf{v}{_{\text{o}}}
\end{align*}
where $\boldsymbol{\tilde{\xi}}_{\text{o}}^{\text{T}}{\mathbf{P}_{\text{o}}}\equiv\mathbf{0}$
is utilized. Therefore 
\begin{align}
\dot{V}_{1,1}= & -\frac{1}{\left\Vert \mathbf{v}{_{\text{o}}}\right\Vert \left\Vert \boldsymbol{\tilde{\xi}}{_{\text{o}}}\right\Vert }\mathbf{v}_{\text{o}}^{\text{T}}{\mathbf{P}_{\text{o}}}\mathbf{v}{_{\text{o}}}\nonumber \\
 & -\frac{{{\kappa}_{{v_{\text{m,1}}}}t}}{\left\Vert \mathbf{v}{_{\text{o}}}\right\Vert \left\Vert \boldsymbol{\tilde{\xi}}{_{\text{o}}}\right\Vert }\biggl(\frac{1}{t}\left(\mathbf{p}{_{\text{o}}}\left(0\right)-\mathbf{p}{_{\text{wp}}}\right)^{\text{T}}{\mathbf{P}_{\text{o}}}\mathbf{v}{_{\text{o}}}\nonumber \\
 & +\mathbf{v}_{\text{o}}^{\text{T}}\mathbf{P}_{\text{o}}\mathbf{v}{}_{\text{o}}\biggl).\label{V2_1}
\end{align}
Since 
\[
\boldsymbol{\dot{\tilde{\xi}}}{_{\text{o}}}=\mathbf{-}\text{sa}{\text{t}}\left(\text{sa}{\text{t}}\left(k_{1}\boldsymbol{\tilde{\xi}}{_{\text{wp}}},v_{\text{m}}\right)-a{_{\text{o}}}\boldsymbol{\tilde{\xi}}{_{\text{o}}},{v_{\text{m}}}\right)-\mathbf{v}{_{\text{o}}}
\]
we have 
\[
\left\Vert \boldsymbol{\dot{\tilde{\xi}}}{_{\text{o}}}\right\Vert \leq{v_{\text{m}}}+\left\Vert \mathbf{v}{_{\text{o}}}\right\Vert .
\]
Then 
\begin{align*}
\left\Vert \boldsymbol{\tilde{\xi}}{_{\text{o}}}\right\Vert  & \leq\left\Vert \boldsymbol{\tilde{\xi}}{_{\text{o}}}\left(0\right)\right\Vert +t\left({v_{\text{m}}}+\left\Vert \mathbf{v}{_{\text{o}}}\right\Vert \right)\\
 & \Rightarrow\\
0 & <\frac{t}{\left\Vert \boldsymbol{\tilde{\xi}}{_{\text{o}}}\left(0\right)\right\Vert +t\left({v_{\text{m}}}+\left\Vert \mathbf{v}{_{\text{o}}}\right\Vert \right)}\leq\frac{t}{\left\Vert \boldsymbol{\tilde{\xi}}{_{\text{o}}}\right\Vert }.
\end{align*}
Therefore, there exists a $t_{1}^{\prime}>0$ such that 
\[
-\frac{t}{\left\Vert \boldsymbol{\tilde{\xi}}{_{\text{o}}}\right\Vert }\leq-\frac{1}{2\left({v_{\text{m}}}+\left\Vert \mathbf{v}{_{\text{o}}}\right\Vert \right)},t\geq t_{1}^{\prime}.
\]
As a result, (\ref{V2_1}) becomes 
\begin{align}
\dot{V}_{1,1} & \leq-c_{1}\mathbf{v}_{\text{o}}^{\text{T}}{\mathbf{P}_{\text{o}}}\mathbf{v}{_{\text{o}}}+\frac{1}{t}c_{0}\left\Vert {\mathbf{P}_{\text{o}}}\mathbf{v}{_{\text{o}}}\right\Vert \nonumber \\
 & =-c_{1}\left\Vert {\mathbf{P}_{\text{o}}}\mathbf{v}{_{\text{o}}}\right\Vert ^{2}+\frac{1}{t}c_{0}\left\Vert {\mathbf{P}_{\text{o}}}\mathbf{v}{_{\text{o}}}\right\Vert \label{dV11}
\end{align}
where 
\begin{align*}
c_{0} & =\frac{\left\Vert \mathbf{p}{_{\text{o}}}\left(0\right)-\mathbf{p}{_{\text{wp}}}\right\Vert }{2\left\Vert \mathbf{v}{_{\text{o}}}\right\Vert \left({v_{\text{m}}}+\left\Vert \mathbf{v}{_{\text{o}}}\right\Vert \right)}\\
c_{1} & =\frac{1}{\left\Vert \mathbf{v}{_{\text{o}}}\right\Vert \left\Vert \boldsymbol{\tilde{\xi}}{_{\text{o}}}\right\Vert }+\frac{{{\kappa}_{{v_{\text{m,1}}}}}}{2\left\Vert \mathbf{v}{_{\text{o}}}\right\Vert \left({v_{\text{m}}}+\left\Vert \mathbf{v}{_{\text{o}}}\right\Vert \right)}.
\end{align*}
\item \textbf{Step 3. Stability analysis. }The term $\mathbf{v}_{\text{o}}^{\text{T}}{\mathbf{P}_{\text{o}}}\mathbf{v}{_{\text{o}}}$
is rewritten as 
\begin{align*}
\mathbf{v}_{\text{o}}^{\text{T}}{\mathbf{P}_{\text{o}}}\mathbf{v}{_{\text{o}}} & =\mathbf{v}_{\text{o}}^{\text{T}}\mathbf{v}{_{\text{o}}}-\frac{1}{\left\Vert \boldsymbol{\tilde{\xi}}{_{\text{o}}}\right\Vert ^{2}}\mathbf{v}_{\text{o}}^{\text{T}}\boldsymbol{\tilde{\xi}}{_{\text{o}}}\boldsymbol{\tilde{\xi}}_{\text{o}}^{\text{T}}\mathbf{v}{_{\text{o}}}\\
 & =\mathbf{v}_{\text{o}}^{\text{T}}\mathbf{v}{_{\text{o}}}\sin^{2}\theta\\
 & =4\mathbf{v}_{\text{o}}^{\text{T}}\mathbf{v}{_{\text{o}}}\sin^{2}\frac{\theta}{2}\cos^{2}\frac{\theta}{2}.
\end{align*}
Since $\theta\in\left[0,\pi\right],$ $\sin\frac{\theta}{2}\cos\frac{\theta}{2}\geq0.$
Then 
\begin{align*}
\left\Vert {\mathbf{P}_{\text{o}}}\mathbf{v}{_{\text{o}}}\right\Vert  & =\sqrt{\mathbf{v}_{\text{o}}^{\text{T}}{\mathbf{P}_{\text{o}}\mathbf{P}_{\text{o}}}\mathbf{v}{_{\text{o}}}}=\sqrt{\mathbf{v}_{\text{o}}^{\text{T}}{\mathbf{P}_{\text{o}}}\mathbf{v}{_{\text{o}}}}\\
 & =2\left\Vert \mathbf{v}{_{\text{o}}}\right\Vert \sin\frac{\theta}{2}\cos\frac{\theta}{2}.
\end{align*}
As a result, (\ref{dV11}) becomes 
\[
\dot{V}_{1,1}\leq-W\left(\cos\frac{\theta}{2}\right).
\]
where 
\begin{align*}
W\left(\cos\frac{\theta}{2}\right) & =4c_{1}\left\Vert \mathbf{v}{_{\text{o}}}\right\Vert ^{2}\sin^{2}\frac{\theta}{2}\cos^{2}\frac{\theta}{2}\\
 & -\frac{2}{t}c_{0}\left\Vert \mathbf{v}{_{\text{o}}}\right\Vert \sin\frac{\theta}{2}\cos\frac{\theta}{2}.
\end{align*}
There are only two solutions to make $W\left(\cos\frac{\theta}{2}\right)=0,$
namely $\theta=0,$ $\pi.$ When $\theta=0,$ the Lyapunov function
$V_{1,1}$ reaches its maximum, namely $V_{1,1}=2.$ When $\theta=\pi,$
the Lyapunov function reach its minimum, namely $V_{1,1}=0.$ First,
we will show that $\theta=0$ is an unstable equilibrium. Given a
perturbation $\Delta\theta>0$ (it is noted that $\theta\in\left[0,\pi\right]$)$,$
there exists a sufficiently large $t_{1}>0$ such that 
\[
\sin\Delta\theta=2\sin\frac{\Delta\theta}{2}\cos\frac{\Delta\theta}{2}>\frac{c_{0}}{tc_{1}\left\Vert \mathbf{v}{_{\text{o}}}\right\Vert },t>t_{1}.
\]
Then 
\[
W\left(\cos\frac{\Delta\theta}{2}\right)<0.
\]
This implies that $V_{1,1}$ will be decreased.$\ $Therefore, $\theta=0$
is an unstable equilibrium. If 
\[
\sin\theta=2\sin\frac{\theta}{2}\cos\frac{\theta}{2}>\frac{c_{0}}{tc_{1}\left\Vert \mathbf{v}{_{\text{o}}}\right\Vert }.
\]
Then 
\[
W\left(\cos\frac{\theta}{2}\right)<0.
\]
It is noticed that $\frac{c_{0}}{tc_{1}\left\Vert \mathbf{v}{_{\text{o}}}\right\Vert }\rightarrow0$
as $t\rightarrow\infty$. Similar to \cite[Th 4.18]{Khalil(1996)},
we can get $\sin\theta\rightarrow0$ as $t\rightarrow\infty$. This
implies $\underset{t\rightarrow\infty}{\lim}\theta\left(t\right)=\pi.$
Only $\theta=\pi$ is a stable equilibrium. Even if the multicopter
only can be stable in one dimensional space when $\theta=0$, which
the measure is 0 on a 2D space. Therefore, $\underset{t\rightarrow\infty}{\lim}\theta\left(t\right)=\pi{\ }${for}
{almost} all $\boldsymbol{\tilde{\xi}}_{\text{o}}\left(0\right).$ 
\end{itemize}

\subsection{Proof of Lemma 4}

\textit{Proof}. Since $\mathbf{X\ }$is a symmetric matrix with rank$\left(\mathbf{X}\right)\leq n-1,$
the matrix $\mathbf{X}$ can be written as 
\[
\mathbf{X=V}^{-1}\text{diag}\left(\begin{array}{cccc}
\lambda_{1} & \cdots & \lambda_{n-1} & 0\end{array}\right)\mathbf{V}
\]
where $\mathbf{V}\in
\mathbb{R}
^{n\times n}$ is the modal matrix, namely its columns are the eigenvectors of $\mathbf{X}$
corresponding to eigenvalues $\lambda_{1},\cdots,\lambda_{n-1},\ 0.$
Then 
\begin{align*}
\rho\mathbf{I}_{n}+\mathbf{X} & =\rho\mathbf{I}_{n}+\mathbf{V}^{-1}\text{diag}\left(\begin{array}{cccc}
\lambda_{1} & \cdots & \lambda_{n-1} & 0\end{array}\right)\mathbf{V}\\
 & =\mathbf{V}^{-1}\text{diag}\left(\begin{array}{cccc}
\rho+\lambda_{1} & \cdots & \rho+\lambda_{n-1} & \rho\end{array}\right)\mathbf{V}.
\end{align*}
Therefore, $\rho$ is an eigenvalue of matrix $\rho\mathbf{I}_{n}+\mathbf{X}.$
Consequently, $\lambda_{\max}\left(\rho\mathbf{I}_{n}+\mathbf{X}\right)\geq\rho>0.$
There exists a vector $\mathbf{x}_{1}\in
\mathbb{R}
^{n}$ with $\left\Vert \mathbf{x}_{1}\right\Vert =1$ such that \cite[p. 176, Theorem (Rayleigh-Ritz)]{Horn(2012)}
\begin{align*}
\lambda_{\max}\left(\rho\mathbf{I}_{n}+\mathbf{X}\right) & =\underset{\left\Vert \mathbf{x}\right\Vert =1}{\max}\mathbf{x}^{\text{T}}\left(\rho\mathbf{I}_{n}+\mathbf{X}\right)\mathbf{x}\\
 & =\mathbf{x}_{1}^{\text{T}}\left(\rho\mathbf{I}_{n}+\mathbf{X}\right)\mathbf{x}_{1}\geq\rho>0.
\end{align*}
Then 
\begin{align*}
\lambda_{\max}\left(\Lambda+\mathbf{X}\right) & =\underset{\left\Vert \mathbf{x}\right\Vert =1}{\max}\mathbf{x}^{\text{T}}\left(\Lambda+\mathbf{X}\right)\mathbf{x}\\
 & \geq\mathbf{x}_{1}^{\text{T}}\left(\Lambda+\mathbf{X}\right)\mathbf{x}_{1}\\
 & =\mathbf{x}_{1}^{\text{T}}\left(\Lambda-\rho\mathbf{I}_{n}\right)\mathbf{x}_{1}+\lambda_{\max}\left(\rho\mathbf{I}_{n}+\mathbf{X}\right).
\end{align*}
Since $\Lambda>\rho\mathbf{I}_{n},$ we have $\mathbf{x}_{1}^{\text{T}}\left(\Lambda-\rho\mathbf{I}_{n}\right)\mathbf{x}_{1}>0.$
Consequently, $\lambda_{\max}\left(\Lambda+\mathbf{X}\right)>\rho>0.$

\bigskip{}

\begin{IEEEbiography}[{\includegraphics[clip,width=1in,height=1.25in]{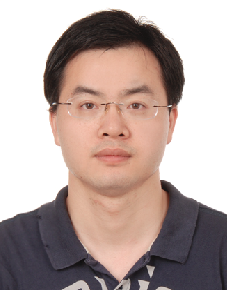}}]{Quan Quan}
received the B.S. and Ph.D. degrees in control science and engineering
from Beihang University, Beijing, China, in 2004 and 2010, respectively.
He has been an Associate Professor with Beihang University since 2013,
where he is currently with the School of Automation Science and Electrical
Engineering. His research interests include reliable flight control,
vision-based navigation, repetitive learning control, and timedelay
systems. 
\end{IEEEbiography}

\begin{IEEEbiography}[{\includegraphics[clip,width=1in,height=1.25in]{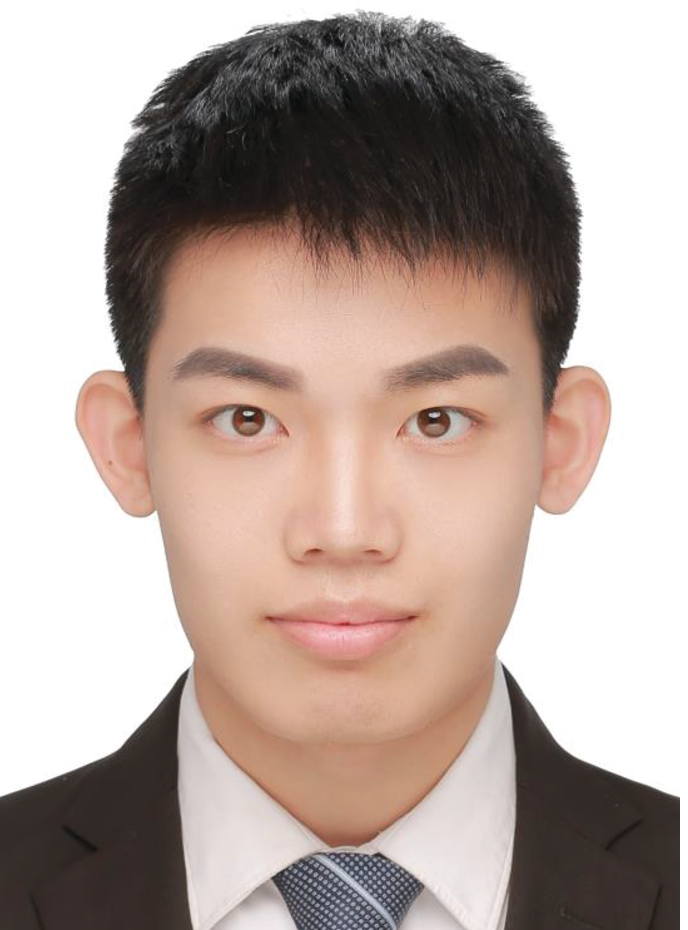}}]{Rao Fu}
received the B.S. degree in control science and engineering from
Beihang University, Beijing, China, in 2017. He is working toward
to the Ph.D. degree at the School of Automation Science and Electrical
Engineering, Beihang University (formerly Beijing University of Aeronautics
and Astronautics), Beijing, China. His main research interests include
UAV traffic control and swarm. 
\end{IEEEbiography}

\begin{IEEEbiography}[{\includegraphics[clip,width=1in,height=1.25in]{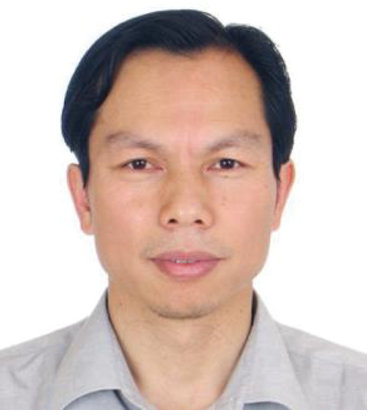}}]{Kai-Yuan Cai}
Kai-Yuan Cai received the B.S., M.S., and Ph.D. degrees in control
science and engineering from Beihang University (Beijing University
of Aeronautics and Astronautics), Beijing, China, in 1984, 1987, and
1991, respectively. He has been a Full Professor with Beihang University
since 1995. He is a Cheung Kong Scholar (Chair Professor), appointed
by the Ministry of Education of China in 1999. His main research interests
include software testing, software reliability, reliable flight control,
and software cybernetics. 
\end{IEEEbiography}


\begin{thebibliography}{10}
\bibitem{IoD(2016)}M. Gharibi, R. Boutaba, S.L. Waslander, ``Internet
of drones'', \emph{IEEE Access}, vol. 4, pp. 1148-1162, 2016.

\bibitem{Devasia(2016)}S. Devasia and A. Lee, ``Scalable low-cost
unmanned-aerial-vehicle traffic network'', \emph{Journal of Air Transportation},
vol. 24, no. 3, pp. 74-83, 2016.

\bibitem{Jenie(2017)}Y. I. Jenie, E. van Kampen, J. Ellerbroek, J.
M. Hoekstra, ``Taxonomy of Conflict Detection and Resolution Approaches
for Unmanned Aerial Vehicle in an Integrated Airspace'', \emph{IEEE
Transactions on Intelligent Transportation Systems}, vol. 18, no.
3, pp. 558-567, 2017.

\bibitem{Mihaela(2019)}M. Mitici, H. A. P. Blom, ``Mathematical
models for air traffic conflict and collision probability estimation'',
\emph{IEEE Transactions on Intelligent Transportation Systems}, vol.
20, no. 3, pp. 1052-1068, 2019.

\bibitem{Kuchar(2000)}J. K. Kuchar and L. C. Yang, ``A review of
conflict detection and resolution modeling methods'', \emph{IEEE
Transactions on Intelligent Transportation Systems}, vol. 1, no. 4,
pp. 179-189, 2000.

\bibitem{Hoekstra2001}J. M. Hoekstra and R. C. J. Ruigrok, R. N.
H. W. Van Gent, ``Free flight in a crowded airspace?'', \emph{Proceedings
of the 3rd USA/Europe Air Traffic Management R\&D Seminar}, 2001.

\bibitem{Lin(2017)}Y. Lin and S. Saripalli, ``Sampling-based path
planning for UAV collision avoidance'', \emph{IEEE Transactions on
Intelligent Transportation Systems}, vol. 18, no. 11, pp. 3179-3192,
2017.

\bibitem{Jenie(2018)}Y. I. Jenie, E. van Kampen, J. Ellerbroek, J.
M. Hoekstra, ``Safety Assessment of a UAV CD\&R System in High Density
Airspace Using Monte Carlo Simulations'', \emph{IEEE Transactions
on Intelligent Transportation Systems}, vol. 19, no. 8, pp. 2686-2695,
2018.

\bibitem{Huang(2019)}S. Huang, R. S. H. Teo, K. K. Tan, ``Collision
avoidance of multi unmanned aerial vehicles: A review'', \emph{Annual
Reviews in Control}, vol. 48, pp. 147-164, 2019.

\bibitem{Mcfadyen(2016)}A. Mcfadyen and L. Mejias, ``A survey of
autonomous vision-based see and avoid for unmanned aircraft systems'',
\emph{Progress in Aerospace Sciences}, vol. 80, pp. 1-17, 2016.

\bibitem{Mueller(2016)}E. R. Mueller and M. Kochenderfer, ``Simulation
comparison of collision avoidance algorithms for small multi-rotor
aircraft'', \emph{AIAA Modeling and Simulation Technologies Conference},
pp. 3674, 2016.

\bibitem{Fiorini(1998)}P. Fiorini and Z. Shiller, ``Motion planning
in dynamic environments using velocity obstacles'', \emph{The International
Journal of Robotics Research}, vol. 17, no. 7, pp. 760-772, 1998.

\bibitem{Thanh(2018)}H. L. N. N. Thanh and S. K. Hong, ``Completion
of collision avoidance control algorithm for multicopters based on
geometrical constraints'', \emph{IEEE Access}, vol. 6, pp. 27111-27126,
2018.

\bibitem{Jenie(2015)} Y. I. Jenie, E.-J. van Kampen, C. C. de Visser,
J. Ellerbroek, J. M. Hoekstra, ``Selective velocity obstacle method
for deconflicting maneuvers applied to unmanned aerial vehicles'',
\emph{Journal of Guidance, Control, and Dynamics}, vol. 38, no. 6,
pp. 1140-1146, 2015.

\bibitem{Jenie(2016)}Y. I. Jenie, E.-J. van Kampen, C. C. de Visser,
J. Ellerbroek, J. M. Hoekstra, ``Three-dimensional velocity obstacle
method for uncoordinated avoidance maneuvers of unmanned aerial vehicles'',
\emph{Journal of Guidance, Control, and Dynamics}, vol. 39, no. 10,
pp. 2312-2323, 2016.

\bibitem{Ong(2016)}H. Y. Ong and M. J. Kochenderfer, ``Markov decision
process-based distributed conflict resolution for drone air traffic
management'', \emph{Journal of Guidance, Control, and Dynamics},
vol. 40, no. 1, pp. 69-80, Oct. 2016.

\bibitem{Saunders(2005)}J. Saunders and B. Call, A. Curtis, R. Beard,
T. McLain, ``Static and dynamic obstacle avoidance in miniature air
vehicles'', \emph{AIAA InfotechAerospace}, pp. 1-14, 2005.

\bibitem{LaValle(2001)}S.M. LaValle and J.J. Kuffner, ``Randomized
kinodynamic planning'',\emph{ International Journal of Robotics Research},
vol. 20, no. 5, pp. 378-400, 2001.

\bibitem{Balazs(2018)}B. Balázs and G. Vásárhelyi, ``Coordinated
dense aerial traffic with self-driving drones'', \emph{2018 IEEE
International Conference on Robotics and Automation (ICRA)}, pp. 6365-6372,
2018.

\bibitem{Viragh(2016)}C. Virágh, M. Nagy, C. Gershenson, G. Vásárhelyi,
``Self-organized UAV traffic in realistic environments'', \emph{2016
IEEE/RSJ International Conference on Intelligent Robots and Systems
(IROS)}, pp. 1645-1652, 2016.

\bibitem{Hernandez(2011)}E. G. Hernandez-Martinez and E. Aranda-Bricaire,
``Convergence and collision avoidance in formation control: A survey
of the artificial potential functions approach'', \emph{Multi-Agent
Systems---Modeling Control Programming Simulations and Applications},
pp. 103-126, 2011.

\bibitem{Boivin(2008)}E. Boivin, A. Desbiens and E. Gagnon, ``UAV
collision avoidance using cooperative predictive control'', \emph{2008
16th Mediterranean Conference on Control and Automation}, pp. 682-688,
2008.

\bibitem{Yang(2018)}X. Yang and P. Wei, ``Autonomous on-demand free
flight operations in urban air mobility using Monte Carlo tree search'',
\emph{2018 8th International Conference for Research in Air Transportation},
2018.

\bibitem{Richards(2002)}A. Richards and J. P. How, ``Aircraft trajectory
planning with collision avoidance using mixed integer linear programming'',
\emph{2002 American Control Conference (IEEE Cat. No. CH37301)}, vol.
3, pp. 1936-1941, 2002.

\bibitem{Alonso(2010)}A. Alonso-Ayuso, L. F. Escudero, F. J. Martín-Campo,
``Collision avoidance in air traffic management: A mixed-integer
linear optimization approach'',\emph{ IEEE Transactions on Intelligent
Transportation Systems}, vol. 12, no. 1, pp. 47-57, 2011.

\bibitem{Lin(2015)}Y. Lin and S. Saripalli, ``Collision avoidance
for UAVs using reachable sets'', \emph{2015 International Conference
on Unmanned Aircraft Systems (ICUAS)}, pp. 226-235, 2015.

\bibitem{Yadollah(2017)}Y. Rasekhipour, A. Khajepour, S. Chen, B.
Litkouhi, ``A potential field-based model predictive path-planning
controller for autonomous road vehicles'', \emph{IEEE Transactions
on Intelligent Transportation Systems}, vol. 18, no. 5, pp. 1255-1267,
2017.

\bibitem{Quan(2017)}Q. Quan, \emph{Introduction to Multicopter Design
and Control}, Singapore:Springer, 2017.

\bibitem{Panagou(2016)}D. Panagou, D. M. Stipanovi\'{c}, P. G. Voulgaris,
``Distributed coordination control for multi-robot networks using
Lyapunov-like barrier functions'', \emph{IEEE Transactions on Automatic
Control}, vol. 61, no. 3, pp. 617-632, 2016.

\bibitem{David(2011)}D. Šišlák, P. Volf, M. P\v{e}chou\v{c}ek, ``Agent-based
cooperative decentralized airplane-collision avoidance'', \emph{ IEEE
Transactions on Intelligent Transportation Systems}, vol. 12, no.
1, pp. 36-46, 2011.

\bibitem{A3}DJI, ``Hardware Introduction''{[}Online{]}, Available:\url{https://developer.dji.com/onboard-sdk/documentation/introduction/osdk-hardware-introduction.html}

\bibitem{Slotine(1991)}J.-J. E. Slotine and W. Li, \emph{Applied
Nonlinear Control}, Englewood Cliffs, NJ:Prentice Hall, 1991.

\bibitem{Thomas(2009)}G. B. Thomas, M. D. Weir, J. Hass, C. Heil,
\emph{Thomas' Calculus}, Boston, MA, USA:Pearson, 2014.

\bibitem{Khalil(1996)}H. K. Khalil, \emph{Nonlinear Systems}, New
Jersey, NJ, USA:Prentice-Hall, vol. 3, 1996.

\bibitem{Horn(2012)}R. A. Horn and C. R. Johnson, \emph{Matrix Analysis},
Cambridge, U.K.:Cambridge University Press, 2012. 
\end{thebibliography}
\end{document}